\documentclass[sn-apa, iicol]{sn-jnl}

\usepackage{bm}
\usepackage{fontawesome}
\usepackage{algorithm}
\usepackage{algpseudocode}
\usepackage{etoolbox}
\usepackage{xcolor}
\usepackage{xspace}

\jyear{2021}%

\theoremstyle{thmstyleone}%
\theoremstyle{thmstyletwo}%

\theoremstyle{thmstylethree}%

\makeatletter
\DeclareRobustCommand\onedot{\futurelet\@let@token\@onedot}
\def\@onedot{\ifx\@let@token.\else.\null\fi\xspace}

\def\eg{\emph{e.g}\onedot} 
\def\ie{\emph{i.e}\onedot} 
 
\def\etc{\emph{etc}\onedot}

\makeatother

\usepackage[capitalize]{cleveref}
\crefname{section}{Sec.}{Secs.}
\Crefname{section}{Section}{Sections}
\Crefname{table}{Table}{Tables}
\crefname{table}{Tab.}{Tabs.}

\newcommand{\Figref}[1]{Figure~\ref{fig:#1}}

\newcommand{\eqnref}[1]{Eq.~\ref{eq:#1}}

\newcommand{\vect}[1]{{{\bf{#1}}}}
\newcommand{\mat}[1]{{{\bf{#1}}}}
\newcommand{\func}[1]{{{\bm{#1}}}}

\DeclareMathOperator*{\argmin}{argmin}
\makeatletter
\patchcmd{\NAT@citex}
  {\@citea\NAT@hyper@{%
     \NAT@nmfmt{\NAT@nm}%
     \hyper@natlinkbreak{\NAT@aysep\NAT@spacechar}{\@citeb\@extra@b@citeb}%
     \NAT@date}}
  {\@citea\NAT@nmfmt{\NAT@nm}%
  \NAT@aysep\NAT@spacechar\NAT@hyper@{\NAT@date}}{}{}

\patchcmd{\NAT@citex}
  {\@citea\NAT@hyper@{%
     \NAT@nmfmt{\NAT@nm}%
     \hyper@natlinkbreak{\NAT@spacechar\NAT@@open\if*#1*\else#1\NAT@spacechar\fi}%
      {\@citeb\@extra@b@citeb}%
     \NAT@date}}
  {\@citea\NAT@nmfmt{\NAT@nm}%
  \NAT@spacechar\NAT@@open\if*#1*\else#1\NAT@spacechar\fi\NAT@hyper@{\NAT@date}}
  {}{}

\makeatother

\begin{document}

\title[Article Title]{Human Trajectory Forecasting with Explainable Behavioral Uncertainty}


\author[1]{\fnm{Jiangbei} \sur{Yue}}
\author[2]{\fnm{Dinesh} \sur{Manocha}}
\author*[1]{\fnm{He} \sur{Wang}}\email{h.e.wang@leeds.ac.uk}


\affil*[1]{\orgname{University of Leeds}, \orgaddress{\street{Woodhouse Lane}, \city{Leeds}, \country{UK}}}
\affil[2]{\orgname{University of Maryland at College Park}, \orgaddress{\city{College Park}, \state{Maryland}, \country{US}}}




\abstract{Human trajectory forecasting helps to understand and predict human behaviors, enabling applications from social robots to self-driving cars, and therefore has been heavily investigated. Most existing methods can be divided into model-free and model-based methods. Model-free methods offer superior prediction accuracy but lack explainability, while model-based methods provide explainability but cannot predict well. Combining both methodologies, we propose a new Bayesian Neural Stochastic Differential Equation model BNSP-SFM, where a behavior SDE model is combined with Bayesian neural networks (BNNs). While the NNs provide superior predictive power, the SDE offers strong explainability with quantifiable uncertainty in behavior and observation. We show that BNSP-SFM achieves up to a 50\% improvement in prediction accuracy, compared with 11 state-of-the-art methods. BNSP-SFM also generalizes better to drastically different scenes with different environments and crowd densities ($\sim$20 times higher than the testing data). Finally, BNSP-SFM can provide predictions with confidence to better explain potential causes of behaviors. The code will be released upon acceptance.}




\maketitle

\section{Introduction}\label{sec1}
Accurate human trajectory forecasting benefits many applications, \eg, social robots, self-driving vehicles, etc~\citep{bennewitz2002learning,thrun2005probabilistic}, and therefore has been studied in areas from computer science, physics, and mathematics to robotics and transportation~\citep{bendali2021recent}. Existing research largely falls into model-free and model-based methods. Model-free methods enjoy the strong data-fitting capacity of data-driven models such as statistical machine learning models \citep{wang_trending_2016,wang2016globally} and deep neural networks (DNNs)~\citep{hochreiter1997long,goodfellow2020generative,kingma2019introduction}. While they provide excellent prediction accuracy, their black-box nature makes it difficult for humans to interpret the learned underlying function. Comparatively, model-based methods are based on explicit systems parameterized as ordinary/partial/stochastic differentiable equations (O/P/SDEs)~\citep{dietrich2021learning} or rule-based systems~\citep{Helbing_Social_1995}. These models are explainable but less accurate in prediction~\citep{Yue_trajectory_2022}, as they do not benefit from training on data (or only on small amounts of data) and therefore are better fit in small data regime. Overall, there seems to be a trade-off between the ability to explain (\ie explainability) and the ability to predict (\ie foresight).

A very recent effort has been focused on eliminating the explainability-foresight trade-off via a hybrid approach~\citep{hossain22sfmgnet,kreiss2021deep,Yue_trajectory_2022}. One can combine a PDE with DNNs for trajectory prediction, where the PDE can explain the behavior while the DNNs can provide data-fitting capacity. However, these approaches either do not capture uncertainty or do not explore the structure of uncertainty, only reporting the average/best performance in prediction without giving the confidence of the prediction~\citep{alahi2016social,gupta2018social}. Generally, this is not ideal because uncertainty modeling is crucial in many tasks, especially in risk-related decision-making~\citep{huang2019uncertainty}. Specific to human trajectories, uncertainty can also help explain behaviors better, \eg why a pedestrian suddenly steers. 

To fill this gap while retaining advantages of hybrid methods, we extend our previous work NSP-SFM~\citep{Yue_trajectory_2022} to propose a new Bayesian Neural Stochastic Differentiable Equation model for trajectory forecasting. Unlike NSP-SFM which learns a \textit{deterministic} steering behavior plus an unexplainable randomness, we further dissect the randomness into \textit{aleatoric uncertainty} caused by behavioral randomness and \textit{epistemic uncertainty} by unobserved reasons~\citep{He_Informative_2020}. We argue that the aleatoric uncertainty should be explainable and the epistemic uncertainty can be unexplainable, because the former is observed in data and can be explained \eg, due to collision avoidance, while the latter is unknown. To this end, we propose to model the steering behavior with the aleatoric uncertainty using a learnable stochastic differentiable equation and learn the epistemic uncertainty with neural nets, leading to a neural stochastic differentiable equation model. Furthermore, to quantify the level of uncertainty in the aleatoric uncertainty, we propose a Bayesian treatment on its attributed factors. 

Specifically, we model the dynamics of a pedestrian trajectory as a second-order SDE, part of which modeling a learnable aleatoric uncertainty caused by stochastic social interactions among pedestrians and stochastic interactions between a pedestrian and the environment. In addition, to account for the epistemic uncertainty, we employ a conditional variational autoencoder (CVAE) that can predict the residual between the prediction from the SDE and the observation~\citep{Yue_trajectory_2022}. Furthermore, to quantify the randomness of the aleatoric uncertainty, we impose priors on the associated factors and learn both the motion dynamics and uncertainty via Bayesian neural networks. We call our model Bayesian Neural Social Physics (BNSP).


We show that our BNSP model can achieve the state-of-the-art performance in standard trajectory prediction tasks across several public datasets and metrics. In addition, the BNSP model can provide not only plausible explanations for predicted trajectories, but also a confidence analysis for the prediction and explanation. Moreover, we demonstrate the better generalizability to unseen scenarios of our BNSP model compared with existing methods. Formally, our contributions include: 
\begin{itemize}
    \item A new Bayesian neural stochastic differentiable equation model for trajectory forecasting and behavior analysis with fine-grained explainable uncertainties;
    \item A new way to combine SDE with neural networks for human trajectory prediction;
    \item A demonstration of the effectiveness of our model over a wide range of tasks and data: trajectory forecasting, behavior explanation with aleatoric uncertainty, generalization to unseen scenarios and high data efficiency.
\end{itemize}

\section{Related Work}\label{sec2}
Human trajectory forecasting predicts future human trajectories based on historical and environmental information. Existing methods can be divided into model-based and model-free methods depending on whether it explicitly models human behaviors or not. 

Model-based approaches are based on first-principles and depend on fundamental assumptions of human motions. With these assumptions, motions can be summarized into rules and deterministic systems, described as ordinary/partial differential equations or optimization problems. In such methods, velocity, acceleration and turning can be assumed to be constant~\citep{schneider2013pedestrian} to model the steering. Pedestrians can be treated as particles that are subject to forces caused by social interactions~\citep{Helbing_Social_1995}. In addition to modeling people as entities,  the influence of their affective states is also studied~\citep{luo2008agent}. Later, data-driven model-based methods are introduced to improve data fitting~\citep{ferrer2014behavior,yan2014modeling}. In such a perspective, model parameters are optimized to fit noisy data~\citep{kim2013predicting} and simulate the behaviors. However, model-based methods always suffer from limited flexibility of data fitting or learning capacity, making them incapable of learning from large amounts of data and predicting accurately. In comparison, our model exploits the strong data-fitting ability of deep neural networks to leverage large data fully for accurate prediction.

More recently, model-free methods based on deep learning have dominated human trajectory forecasting by demonstrating their surprisingly high prediction accuracy. Social LSTM~\citep{alahi2016social} uses Long Short-Term Memory (LSTM) networks to model social interactions and learn from temporal data. This work inspires other methods based on recurrent neural networks~\citep{bartoli2018context,vemula2018social,zhang2019sr} and more recent applications of transformers~\citep{yu2020spatio,giuliari2021transformer,wong2022view}. In addition, other deep neural networks have also been explored. Deep generative models such as generative adversarial networks and variational autoencoders are exploited to handle uncertainties in the future trajectories and generate multiple acceptable predictions~\citep{gupta2018social,kosaraju2019social,ivanovic2019trajectron,mangalam2020not,xu2022socialvae}. Graph and temporal convolutional neural networks provide a novel, efficient way to model interactions between people~\citep{mohamed2020social,shi2021sgcn}. Besides high prediction accuracy, it is found that these models have limited explainability and cannot generalize to drastically different scenarios well~\citep{Yue_trajectory_2022}. Different from existing deep learning methods, our model not only provides higher accuracy in prediction but also possesses explainability and better generalizability, benefiting from our embedded explicit model.

Very recently, hybrid approaches combining model-based and model-free methods have emerged. Physical models are embedded into deep neural networks~\citep{hossain22sfmgnet,kreiss2021deep,Yue_trajectory_2022} to balance explainability and prediction accuracy. Our method falls into this category. Compared with these approaches, our model further explores the fine-grained structure of uncertainty, \ie, aleatoric and epistemic uncertainty. As a result, our model can provide more accurate prediction, stronger explainability and better generalization in different scenarios.

At a high level, our method is part of a recent attempt in physics-based deep learning~\citep{karniadakis2021physics} to combine deep neural networks with differential equations. Existing applications include finite element mesh generation~\citep{Zhang_MeshingNet_2020,zhang_meshingnet3d_2021}, reduced-order modeling~\citep{Shen_high_2021}, differentiable simulation~\citep{Gong_finegrained_2022}, \etc. Compared with these methods, our research focuses on human trajectory forecasting.

\section{Methodology}\label{sec3}
We first introduce the background of Bayesian neural stochastic differential equations in \cref{sec:background}. Then we introduce a new Stochastic Social Physics model in \cref{sec:snsp}. Next, we introduce a new Bayesian treatment on our stochastic social physics model to derive our general Bayesian Social Physics Model (BNSP) as a general framework in \cref{sec:BNSP}. Further, we instantiate our BNSP with a stochastic social force model by augmenting a previous deterministic social force method into a stochastic one in \cref{sec:socialForces}. Finally, we derive the inference method in \cref{sec:inference} and provide the implementation details in \cref{sec:implementation}.


\subsection{Background}
\label{sec:background}
Bayesian neural stochastic differential equation~\citep{li2020scalable, opper2019variational,tzen2019neural} is an extension and a combination of both Bayesian neural networks \citep{jospin2022hands} and stochastic differential equations (SDEs) \citep{dietrich2021learning} that allows for the integration of uncertainty quantification and model-based representation of dynamical systems. In this approach, neural networks are used to approximate the solution of a SDE, and Bayesian inference is used to estimate the parameters of these neural networks. This results in a probabilistic model that captures both the dynamics of the system and the uncertainty associated with the model parameters. Bayesian neural stochastic differential equation models have applications in various fields, including finance, physics, and biology. They can be used for tasks such as prediction, parameter estimation, and control of dynamic systems. The ability to model uncertainty in the dynamics of these systems makes these models particularly useful for behavioral analysis under uncertainty.

A SDE is a differential equation that includes a stochastic component and can capture the randomness inherent to dynamical systems. A typical SDE describes a system evolving over time with random fluctuations:
\begin{equation}
\label{eq:t_sde}
\mathrm{d}\func{X}_t = \func{\mu}(t,\func{X}_t)\mathrm{d}t + \func{\sigma}(t,\func{X}_t)\mathrm{d}\func{W}_t,
\end{equation}
where $\func{X}_t$ is a stochastic process of the system state, $\func{\mu}, \func{\sigma}$ are two real-valued functions and $\func{W}(t)$ is a Wiener process. From \cref{eq:t_sde}, we have its integral form:
\begin{equation}
\label{intergral_sde}
\begin{split}
    \func{X}_{t+s}-\func{X}_t=&\int_{t}^{t+s}\func{\mu}(u,\func{X}_u)\mathrm{d}u \\
    +& \int_{t}^{t+s}\func{\sigma}(u,\func{X}_u)\mathrm{d}\func{W}_u. 
\end{split}
\end{equation}
In contrast to ODEs/PDEs, SDEs explicitly consider the randomness whose solutions are stochastic processes. \cref{intergral_sde} demonstrates that the change of the solution $\func{X}_{t}$ is the sum of an ordinary Lebesgue integral of $\func{\mu}$ (the first term) and an It\^{o} integral (the second term). In general, we refer to the function $\mu$ and $\sigma$ as the drift coefficient and the diffusion coefficient, respectively. The solution $\func{X}_{t}$ is called the diffusion processes and satisfies the Markov property. Therefore, we can model the pedestrian motion with uncertainty from the view of SDEs, which naturally captures the uncertainty of future trajectories.

\subsection{Stochastic Neural Social Physics}
\label{sec:snsp}
\textbf{Notation}. Assuming $\vect{p}^t\in \mathbb{R}^2$ is the 2D location of a pedestrian at time $t$, we represent the whole trajectory as a function $\func{p}(t)$. In data, we assume such a trajectory is observed discretely in time $\{ \vect{p}^0, \vect{p}^1, \cdots, \vect{p}^T\}$. Therefore, $\mat{P} = \{ \vect{p}_i^t \}_{i=1:N}^{t=1:T}\in \mathbb{R}^{N\times T \times 2}$ represents a set of trajectories of $N$ pedestrians of length $T$. Given $\vect{p}_i^t$ of the $i$th person, we consider his/her neighborhood set $\mat{\Omega}_i^t$ which includes the indices of other nearby pedestrians, $\vect{p}_j^t$ with $ j \in \mat{\Omega}_i^t$, all of whom influence the motion of pedestrian $i$. The neighborhood is also a function of time $\mat{\Omega(t)}$. 

In Neural Social Physics (NSP)~\citep{Yue_trajectory_2022}, the state of a pedestrian at $t$ is \vect{q}=[$\vect{p}$, $\dot{\vect{p}}$]$^\mathbf{T}$, and the trajectory can be formulated as:

\begin{equation}
\label{eq:ode}
\left\{
\begin{aligned}
    \mathrm{d}\func{p}(t) =& \func{\dot{p}}(t)\mathrm{d}t + \alpha(\vect{q}^{t:t-M})  \\ 
    \mathrm{d}\func{\dot{p}}(t) =&  \func{F}_{goal} + \func{F}_{col} + \func{F}_{env},
\end{aligned}
\right.
\end{equation}
where $\func{F}_{goal}$, $\func{F}_{col}$, $\func{F}_{env}$ are \textit{deterministic} forces parameterized by neural networks to model social interactions. $\alpha(\vect{q}^{t:t-M})$ is another neural network to take into account any randomness that has not been captured in $\mathrm{d}\func{\dot{p}}(t)$.  Although NSP shows superior performance, it ignores the fine-grained structure of the randomness in the data and simply captures it in $\alpha(\vect{q}^{t:t-M})$.

We propose to divide the randomness into aleatoric and epistemic uncertainty and learn them separately in trajectory forecasting. This is because these two uncertainties come from different sources and bear different meanings. Aleatoric uncertainty arises from the steering behavior which is random \eg when avoiding other pedestrians or obstacles, while epistemic uncertainty is caused by unknown factors \eg affective state, sensor error, \etc~\citep{He_Informative_2020}. Explicitly capturing the aleatoric uncertainty can help explain the behavior as well as give confidence about the prediction. 

To this end, we model the aleatoric uncertainty as \textit{random} forces rising from social interactions. We first formulate the dynamics of a person (agent) in a crowd as:
\begin{equation}
\label{eq:bde}
\left\{
\begin{aligned}
    \mathrm{d}\func{p}(t) =& \func{\dot{p}}(t)\mathrm{d}t  \\ 
    \mathrm{d}\func{\dot{p}}(t) =&  \func{f}_{\eta, \phi}(t, \func{p}(t), \func{\dot{p}}(t), \mat{\Omega}(t), {\vect{p}^T}, \mat{E})\mathrm{d}t  \\
    + \func{\sigma}_{\eta, \phi}&(t, \func{p}(t), \func{\dot{p}}(t), \mat{\Omega}(t), {\vect{p}^T}, \mat{E})\mathrm{d}\func{W}(t) \\
    \func{p}(0) = &\vect{p}^0, \func{\dot{p}}(0) = \vect{\dot{p}}^0, \func{p}(T) = \vect{p}^T
\end{aligned}
\right.
\end{equation}
given the initial position $\vect{p}^0$, initial velocity $\vect{\dot{p}}^0$ and the destination $\vect{p}^T$. Here $\func{\dot{p}}(t)$ and $\mathrm{d}\func{\dot{p}}(t)$ denote the first-order and the second-order dynamics of positions $\func{p}(t)$. $\func{f}$ and $\func{\sigma}$ are functions governing the dynamics. $\func{W}(t)$ is a Wiener process such that for any $t_2 > t_1$, $\func{W}(t_2) - \func{W}(t_1)$ is normally distributed with mean $0$ and variance $t_2 - t_1$. $\eta$ is the set of explainable parameters and $\phi$ is the set of unexplainable parameters such as neural network weights. $\mat{\Omega}(t)$ represents the neighborhood set. $\mat{E}$ is the environment (\eg,  obstacles). From \eqnref{bde}:
\begin{equation}
\begin{split}
    \bm{p^T} - \bm{p^0} &= \int_{t=0}^T \bm{\dot{p}(t)}\mathrm{d}t \\
    &=\int_{t=0}^T (\int \bm{f} \mathrm{d}t + \int \bm{\sigma} \mathrm{d}\func{W}(t)) \mathrm{d}t. 
\end{split}
\end{equation}
where $\func{f}$ is a deterministic steering behavior and $\func{\sigma}\Delta{W(t)}$ is the steering randomness. \eqnref{bde} explicitly considers the aleatoric uncertainty, but it alone cannot fully describe the motion. Therefore, we add another term, $\func{\varepsilon}(t, \vect{p}^{t:t-M})$, to model the epistemic uncertainty, which is time-dependent and depends on the brief history $\vect{p}^{t:t-M}$ where $M$ is the length of the history. Therefore, the full model becomes:
\begin{equation}
\label{eq:bde_full}
\left\{
\begin{aligned}
    \mathrm{d}\func{p}(t) =& \func{\dot{p}}(t)\mathrm{d}t + \func{\varepsilon}(t, \vect{p}^{t:t-M}) \\ 
    \mathrm{d}\func{\dot{p}}(t) =&  \func{f}_{\eta, \phi}(t, \func{p}(t), \func{\dot{p}}(t), \mat{\Omega}(t), {\vect{p}^T}, \mat{E})\mathrm{d}t  \\
     + \func{\sigma}_{\eta, \phi}&(t, \func{p}(t), \func{\dot{p}}(t), \mat{\Omega}(t), {\vect{p}^T}, \mat{E})\mathrm{d}\func{W}(t)\\
\end{aligned}
\right.
\end{equation}
where we omit the boundary conditions. To fit this model to discrete observations in time, we discretize \eqnref{bde_full} into:
\begin{align}
\left\{
\begin{aligned}
    {\vect{p}}^{t+\Delta{t}} - {\vect{p}}^t =& {\vect{\dot{p}}}^{t+\Delta{t}}\Delta{t} + \func{\varepsilon}^{t, \vect{p}^{t:t-M}} \\ 
    {\vect{\dot{p}}}^{t+\Delta{t}} - {\vect{\dot{p}}}^t = &{\func{f}}_{\eta, \phi}(t, \func{p}(t), \func{\dot{p}}(t), \mat{\Omega}(t), {\vect{p}^T}, \mat{E})\Delta{t} \nonumber\\
    + {\func{\sigma}}_{\eta, \phi}&(t, \func{p}(t), \func{\dot{p}}(t), \mat{\Omega}(t), {\vect{p}^T}, \mat{E})\Delta{\func{W}(t)} \nonumber.
\end{aligned}
\right.
\end{align}
Finally, the full forward model is:
\begin{equation}
\label{eq:dsde}
    \vect{p}^{t+\Delta{t}} = \vect{p}^t + \vect{\dot{p}}^t\Delta{t} + \func{f}\Delta{t}^2 + \func{\sigma}\Delta{t}\Delta{\func{W}(t)} + \func{\varepsilon}^{t, \vect{p}^{t:t-M}}
\end{equation}
\Cref{eq:dsde} is the main (stochastic) prediction model employed in this paper. It is a general model and can ideally incorporate any $\func{f}$ with a second-order differentiability. We instantiate it later.

\subsection{Bayesian Neural Social Physics}
\label{sec:BNSP}
A key difference between our method and previous ones is we aim to quantify the aleatoric uncertainty, through estimating the explainable parameters $\eta$ in \cref{eq:bde_full} during inference. To this end, we propose a Bayesian treatment on them. Note we temporally ignore unexplainable parameters $\phi$ as they are fixed once learned. Given a new brief history $\vect{\hat{p}}^h=\{ \vect{\hat{p}}^0, \vect{\hat{p}}^1, \cdots, \vect{\hat{p}}^{t_h}\}$, we predict the future trajectory $\vect{\hat{p}}^{f}=\{ \vect{\hat{p}}^{t_h+1}, \vect{\hat{p}}^{t_h+2}, \cdots, \vect{\hat{p}}^{t_h+t_f}\}$ in the testing phase. A Bayesian predictor can be represented as:  
\begin{align}
\label{eq:jointProb}
    p(\vect{P}, &\vect{\hat{p}}^h, \vect{\hat{p}}^f) = \int p(\vect{P}, \vect{\hat{p}}^h, \vect{\hat{p}}^f, \eta)\mathrm{d}\eta = \nonumber\\ 
    &\int p(\vect{\hat{p}}^f \mid \vect{\hat{p}}^h, \mat{P}, \eta) p(\vect{\hat{p}}^h, \mat{P}, \eta) \mathrm{d}\eta = \nonumber\\ 
    &\int p(\vect{\hat{p}}^f \mid \vect{\hat{p}}^h, \eta) p(\eta\mid\mat{P})p(\mat{P})p(\vect{\hat{p}}^h) \mathrm{d}\eta = \nonumber \\
    &\int p(\vect{\hat{p}}^f \mid \vect{\hat{p}}^h, \eta)p(\eta\mid\mat{P})\mathrm{d}\eta,
\end{align}
where $\eta$ is a latent variable. $p(\vect{\hat{p}}^f \mid \vect{\hat{p}}^h, \mat{P}, \eta) = p(\vect{\hat{p}}^f \mid \vect{\hat{p}}^h, \eta)$ because $\mat{P}$ is given and not part of the prediction process. $p(\vect{\hat{p}}^h, \mat{P}, \eta) = p(\eta \mid \vect{\hat{p}}^h, \mat{P})p(\vect{\hat{p}}^h, \mat{P})=p(\eta\mid\mat{P})p(\vect{\hat{p}}^h)p(\mat{P})=p(\eta\mid\mat{P})$ as $\vect{\hat{p}}^h$ is not used for estimating $\eta$, and $\vect{\hat{p}}^h$ and $\mat{P}$ are observed. After learning $p(\eta\mid\mat{P})$, the model predicts via $p(\vect{\hat{p}}^f \mid \vect{\hat{p}}^h, \eta)$ across all $\eta$. With Bayesian inference to learn $p(\eta\mid\mat{P})$, we obtain our full framework BNSP by using \cref{eq:dsde} for $p(\vect{\hat{p}}^f \mid \vect{\hat{p}}^h, \eta)$. BNSP benefits from the quantification of uncertainty during prediction.

\textbf{Remarks.} Most existing deep learning methods can be represented as \cref{eq:jointProb}, where $p(\vect{\hat{p}}^f \mid \vect{\hat{p}}^h, \eta)$ is realized as neural networks (NNs) parameterized by 
 learned parameters~\citep{alahi2016social,gupta2018social,mangalam2021goals}. Deviating from these unexplainable models, recently some research~\citep{hossain22sfmgnet,kreiss2021deep,Yue_trajectory_2022} shows that $p(\vect{\hat{p}}^f \mid \vect{\hat{p}}^h, \eta)$ can be realized as ODEs/PDEs with model parameters and motion randomness captured by NNs. However, while the ODEs/PDEs provide explainability to some extent, they are intrinsically deterministic so that the randomness is either not captured or still captured by additional neural network components hence unexplainable. Their learned randomness does not truly capture the posterior $p(\eta\mid\mat{P})$. Explicitly capturing and explaining such uncertainty is crucial for analysis and prediction, which can be naturally captured by Bayesian inference, as proposed by us.

\subsection{Instantiation of BNSP with Stochastic Social Forces}
\label{sec:socialForces}

\begin{figure}[tb]
\centering
\includegraphics[width=0.9\linewidth]{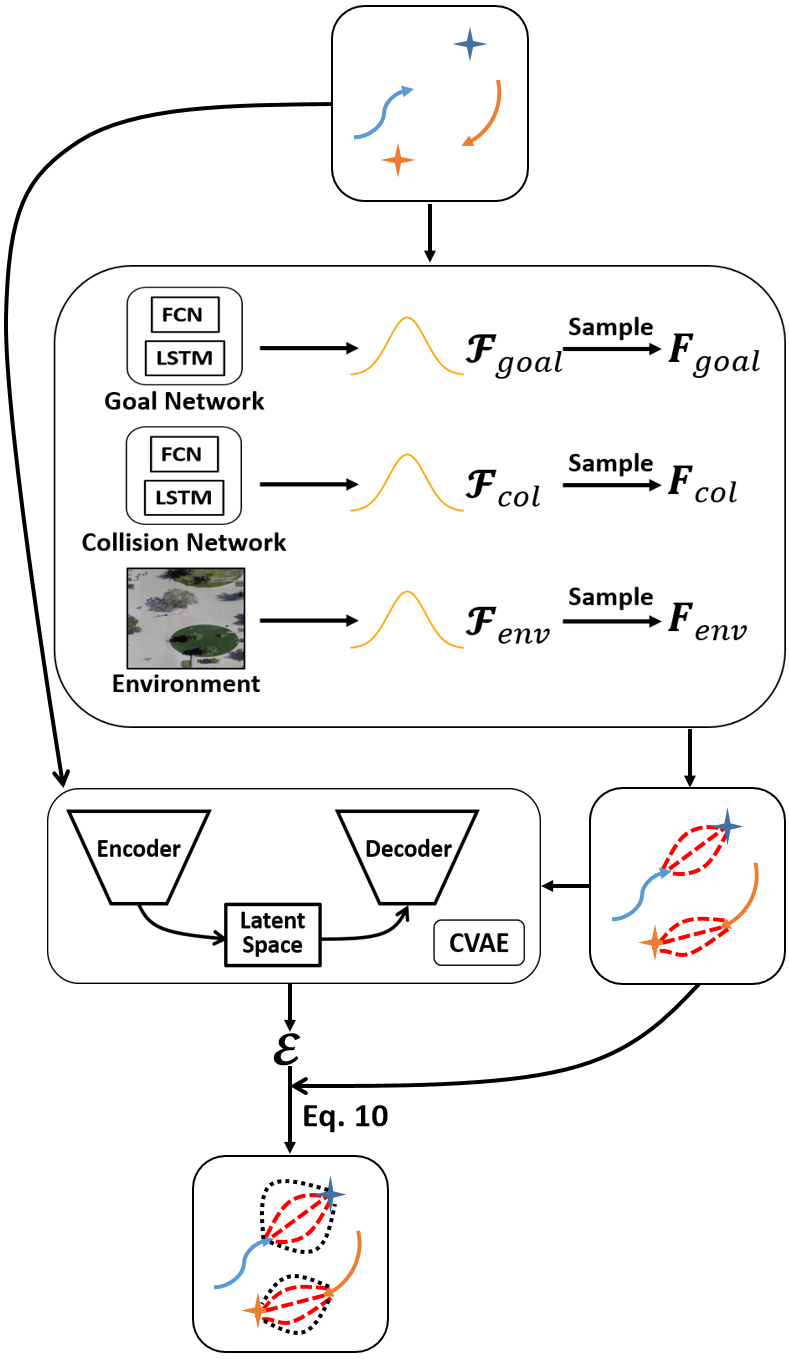}
\caption{\textbf{Overview of BNSP-SFM}. $\func{\mathcal{F}}_{goal}$ and $\func{\mathcal{F}}_{col}$ are estimated at each step through the attraction network and the evasion network. $\func{\mathcal{F}}_{env}$ is a normal distribution related to the environment with the learnable mean and variance. For every step, we sample $\vect{F}^t_{goal}\sim \func{\mathcal{F}}_{goal}$, $\vect{F}^t_{col}\sim\func{\mathcal{F}}_{col}$ and $\vect{F}^t_{env}\sim \func{\mathcal{F}}_{env}$ and sample an observational noise $\func{\varepsilon}$ from the CVAE, then predict the future position via \cref{eq:bnsde}.} 
\label{fig:model}
\end{figure}

Similar to \citep{Yue_trajectory_2022}, we instantiate $\func{f}$ using social forces~\citep{Helbing_Social_1995} but augment it with stochastic forces instead of deterministic ones. One key difference in the modeling choice between our model (\cref{eq:bde_full}) and NSP-SFM~\citep{Yue_trajectory_2022} (\cref{eq:ode}) is that NSP-SFM assumes all the randomness happens at the first-order, while BNSP assumes the epistemic uncertainty is captured at the first order but the aleatoric uncertainty is captured at the second order, which is more sensible as this is where the stochastic forces caused by social interactions are modeled. We name our model BSNP-SFM.

Similar to \cref{eq:ode}, we consider three main factors for the behavior in \cref{eq:dsde}: stochastic goal attraction $\func{\mathcal{F}}_{goal}$, stochastic collision avoidance $\func{\mathcal{F}}_{col}$ and stochastic environment repulsion $\func{\mathcal{F}}_{env}$. To this end, we specify the second-order part of \cref{eq:dsde} as:
\begin{equation}
\label{eq:forces}
    \func{f}+\func{\sigma}\frac{\Delta{W(t)}}{\Delta{t}} = \func{\mathcal{F}}_{goal} + \func{\mathcal{F}}_{col} + \func{\mathcal{F}}_{env},  
\end{equation}

Substituting \cref{eq:forces} into \cref{eq:dsde} gives:
\begin{equation}
\label{eq:bnsde}
\begin{split}
    \vect{p}^{t+\Delta{t}}& =  \vect{p}^t + \vect{\dot{p}}^t\Delta{t} +  \\
    & (\vect{F}^t_{goal} + \vect{F}^t_{col} + \vect{F}^t_{env})\Delta{t}^2 + \func{\varepsilon}^{t,\vect{p}^{t:t-M}} \\
    \vect{F}^t_{goal}& \sim \func{\mathcal{F}}^t_{goal}, \vect{F}^t_{col}\sim\func{\mathcal{F}}^t_{col},  \vect{F}^t_{env}\sim\func{\mathcal{F}}_{env},
\end{split}
\end{equation}
where $\func{\mathcal{F}}_{goal}$ and $\func{\mathcal{F}}_{col}$ are time-varying Gaussians. $\func{\mathcal{F}}_{env}$ is a static Gaussian. The overview of our model is shown in~\Figref{model}. After training, given an input trajectory and an endpoint, we predict distributions $\func{\mathcal{F}}^t_{goal}$,  $\func{\mathcal{F}}^t_{col}$ and $\func{\mathcal{F}}_{env}$ by neural networks; then we can sample $\vect{F}^t_{goal}\sim\func{\mathcal{F}}^t_{goal}$, $\vect{F}^t_{col}\sim\func{\mathcal{F}}^t_{col}$,  $\vect{F}^t_{env}\sim\func{\mathcal{F}}_{env}$ and $\func{\varepsilon}^{t,\vect{p}^{t:t-M}}$ at time $t$. Iteratively, we can predict positions via solving \cref{eq:bnsde}. Note that, in \cref{eq:bde_full}, we assume that destinations $\vect{p}^T$ are given, although they are not available directly during prediction. Therefore, we employ the pre-trained Goal Sampling Network (GSN)~\citep{Yue_trajectory_2022} to sample $\vect{p}^T$ in advance for prediction during testing.  

\subsubsection{Individual Neural Networks}

\begin{figure}[tb]
\centering
\includegraphics[width=\linewidth]{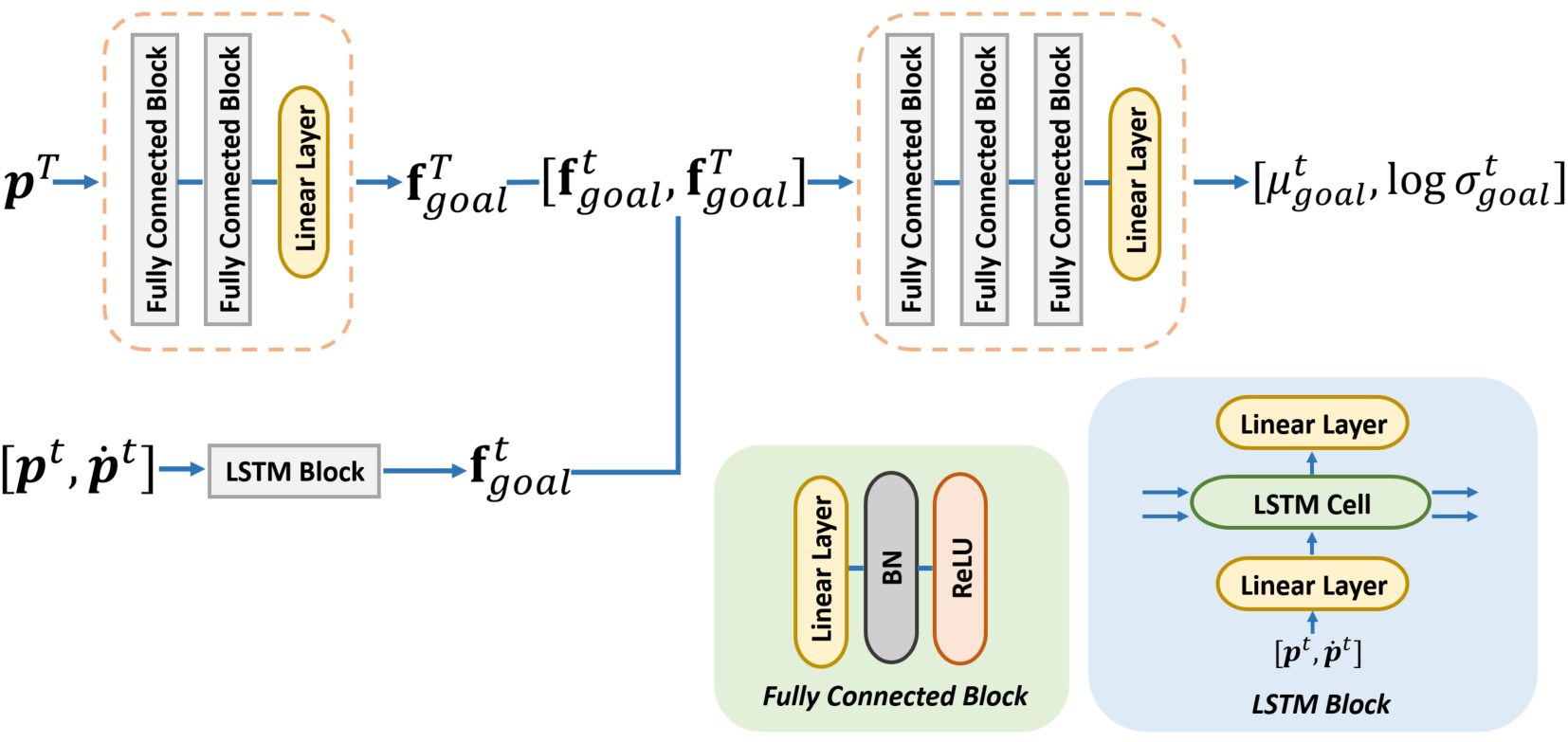}
\caption{\textbf{The structure of the goal network}.}
\label{fig:attraction}
\end{figure}

\textbf{Stochastic Goal Attraction} Humans are constantly attracted to their destinations. However, the magnitude and direction of such an attraction can vary over time and in different circumstances (\eg, detour for collision avoidance). We model such goal attraction as a stochastic force:
\begin{equation}
\label{eq:f_att}
\begin{split}
    &\vect{F}^t_{goal} = (\frac{\vect{p}^T-\vect{p}^t}{(T-t)\Delta{t}} - \vect{\dot{p}}^t)k_{goal}^t\\
    &k_{goal}^t \sim \mathcal{N}_{\phi_1}(\mu_{goal}^t, \sigma_{goal}^{t \, 2}),
\end{split}
\end{equation}
where $(\frac{\vect{p}^T-\vect{p}^t}{(T-t)\Delta{t}} - \vect{\dot{p}}^t)$ is the expected velocity correction on the current velocity $\vect{\dot{p}}^t$ towards $\vect{p}^T$. $\mu_{att}^t$ and $\sigma_{att}^t$ are the mean and standard deviation functions at $t$, realized by the Goal Network (GN) with parameters $\phi_1$:
\begin{align}
\label{eq:attractionNet}
    [\mu_{goal}^t, \sigma_{goal}^t]_{\phi_1}& = GN_{\phi_1}(\vect{p}^t, \vect{\dot{p}}^t, \vect{p}^T)
\end{align}
whose architecture is shown in \cref{fig:attraction}. $[\vect{p}^t, \vect{\dot{p}}^t]$ is encoded and fed into a Long Short-Term Memory (LSTM) network~\citep{hochreiter1997long}. The output of the LSTM is transformed by a linear layer to get the feature $\vect{f}_{goal}^t$. The first orange block (orange dashed lines), which consists of two fully connected blocks and one linear layer, is used to encode the destination $\vect{p}^T$ into $\vect{f}_{goal}^T$. Then the concatenated feature $[\vect{f}_{goal}^t,\vect{f}_{goal}^T]$ is fed into the second orange block to output the distribution parameters $[\mu_{k_{goal}}, \log{\sigma_{k_{goal}}}]$. Instead of learning $\sigma_{k_{goal}}$ directly, we learn its logarithm to ensure that every output dimensions have the same range. The dimensions of each linear layer in the first and second orange block are [64, 256, 16] and [512, 256, 512, 2], respectively. In the LSTM block, the linear layers before and after the LSTM have dimensions 64 and 16, respectively. We use an LSTM with 256 dimensions.

\begin{figure}[tb]
\centering
\includegraphics[width=\linewidth]{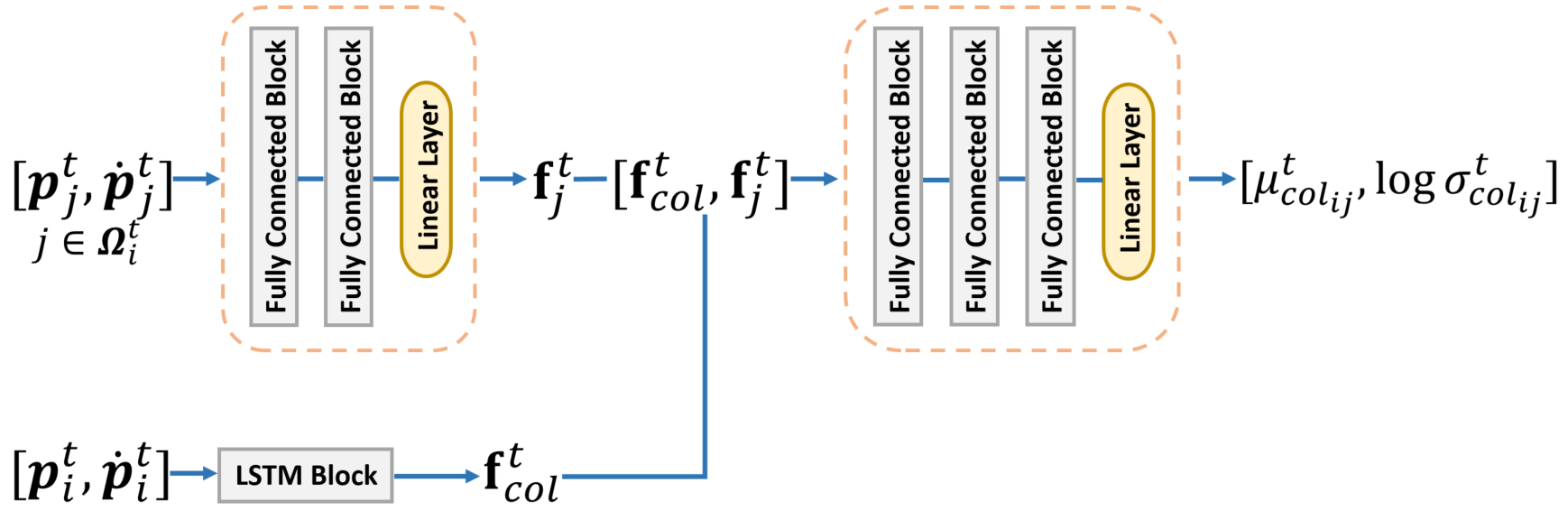}
\caption{\textbf{The structure of the collision network}.}
\label{fig:evasion}
\end{figure}

\textbf{Stochastic Collision Avoidance} Given the $i$th pedestrian at $t$, $\vect{p}_i^t$, its neighborhood $\vect{\Omega}_i^t$ and any pedestrian $\vect{p}_j^t$, $j\in\vect{\Omega}^t_i$, we define the collision avoidance factor $\func{\mathcal{F}}_{col}^t$  as follows:

\begin{equation}
\label{eq:repulsion}
\begin{split}
    &\func{\mathcal{F}}_{col}^t = \sum_{j=0}^m \func{\mathcal{F}}_{col_{ij}}^t, \text { where } j \in \vect{\Omega}_i^t, \vect{F}^t_{col_{ij}}\sim\func{\mathcal{F}}^t_{col_{ij}} \\
    &\vect{F}^{t}_{col_{ij}} =-\nabla_{\vect{r}_{ij}^t}r_{col}e^{-\lVert \vect{r}_{ij}^t\rVert/r_{col}}k^t_{col_{ij}},\\
    &\vect{r}_{ij} = \vect{p}_i^t - \vect{p}_j^t, \text{   } k^t_{col_{ij}} \sim \mathcal{N}_{\phi_2}(\mu^t_{col_{ij}}, \sigma^{t \, 2}_{col_{ij}}),
\end{split}    
\end{equation}
where $\mu^t_{eva_{ij}}$ and $\sigma^t_{eva_{ij}}$ are the mean and standard deviation functions. Similar to~\citep{Helbing_Social_1995}, $r_{col}e^{-\lVert \vect{r}_{ij} \rVert/r_{col}}$ is a repulsive potential energy function with a radius $r_{col}$ (hyperparameter), and the negative gradient w.r.t. $r_{ij}$ gives a repulsive force. However, such repulsion between pedestrians has randomness~\citep{wang_path_2016,He_Informative_2020}. Therefore, our model considers a time-varying Gaussian with learnable mean and variance, so that the collision avoidance factor $\func{\mathcal{F}}_{col}^t$ is a time-varying Gaussian mixture. We realize $\mu_{col_{ij}}$ and $\sigma_{col_{ij}}$ by the Collision Network (CN) parameterized by $\phi_2$:
\begin{align}
\label{eq:repulsionNet}
    [&\mu_{col_{ij}}^t, \sigma_{col_{ij}}^t]_{\phi_2} = CN_{\phi_2}( \vect{p}^t_i, \vect{\dot{p}}^t_i, \vect{p}^t_j, \vect{\dot{p}}^t_j)
\end{align}
whose architecture is shown in \cref{fig:evasion}. For any neighbor $\vect{p}_j^t$ in $\mat{\Omega}_i^t$, the collision network encodes $[\vect{p}_i^t, \vect{\dot{p}}_i^t]$ and $[\vect{p}_j^t, \vect{\dot{p}}_j^t]$ to features $\vect{f}_{col}^t$ and $\vect{f}_j^t$, respectively. The concatenated feature $[\vect{f}_{col}^t, \vect{f}_j^t]$ is fed into a decoder to output the distribution parameters. The collision network has the same architecture and dimensions as the goal network except that the input dimension of the first orange block is 4.

\textbf{Stochastic Environment Repulsion} To model how people avoid obstacles in the environment, given the position $\vect{p}^t$ and an obstacle position $\vect{p}_{obj}$, we define the repulsion as:

\begin{equation}
\label{eq:f_env}
\begin{split}
    &\func{\mathcal{F}}_{env}^t = \sum_{obj\in\mat{E}} \func{\mathcal{F}}_{obj}^t, \vect{F}^t_{obj}\sim\func{\mathcal{F}}^t_{obj}\\
    &\vect{F}_{obj}^t = (\frac{\vect{p}^t - \vect{p}_{obj}}{\lVert \vect{p}^t - \vect{p}_{obj} \rVert^2_2})k_{env}, k_{env} \sim \mathcal{N}(\mu_{obj}, \sigma_{obj}^2)
\end{split}
\end{equation}
where $\mu_{obj}$ and $\sigma_{obj}$ are the mean and standard deviation. Different from \cref{eq:f_att} and \cref{eq:repulsion}, $k_{env}$ is assumed to be time-independent and agent-independent. This is because we observe similar influences of obstacles on different pedestrians. Therefore, unlike $\func{\mathcal{F}}_{att}$ and $\func{\mathcal{F}}_{col}$, we learn $\mu_{obj}$ and $\sigma_{obj}$ directly and do not use neural networks.

\paragraph{Epistemic Uncertainty} 
\begin{figure}[tb]
\centering
\includegraphics[width=\linewidth]{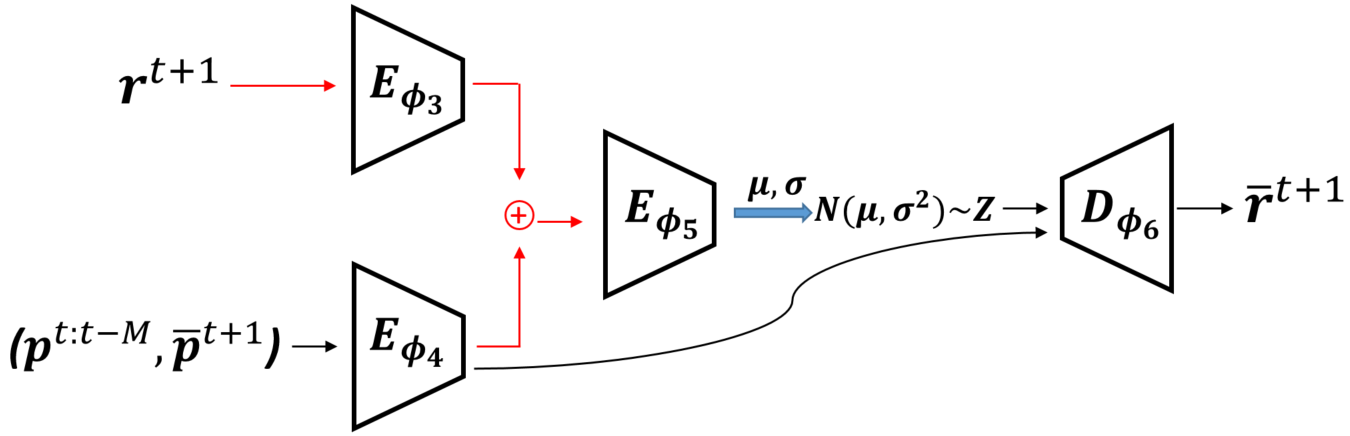}
\caption{\textbf{The Structure of CVAE}. The residual $\vect{r}^{t+1} = \vect{p}^{t+1} -  \vect{\bar{p}}^{t+1}$. $E_{\phi_3}$, $E_{\phi_4}$, $E_{\phi_5}$ and $D_{\phi_6}$ are all multi-layer perceptrons. The red
connections are only used in the training phase.}
\label{fig:cvae}
\end{figure}
Finally, we specify the epistemic uncertainty term $\func{\varepsilon}^{t,\vect{p}^{t:t-M}}$. Here, we employ the same strategy as~\citep{Yue_trajectory_2022} and assume the observation noise has a well-behaved distribution in the latent space, rather than the data space. This is because the epistemic term captures all the residual randomness that is not captured by \cref{eq:bde} and its distribution is arbitrary. Therefore, we use a variational autoencoder to learn its distribution. The architecture of the Conditional Variational Autoencoder (CVAE) is shown in \cref{fig:cvae}. We use the CVAE to reconstruct the residual $\vect{r}^{t+1} = \vect{p}^{t+1} -  \vect{\bar{p}}^{t+1}$ between the ground truth $\vect{p}^{t+1}$ and the prediction $\vect{\bar{p}}^{t+1}$ only considering the aleatoric uncertainty. The residual $\vect{r}^{t+1}$ is encoded by $E_{\phi_3}$, while the history condition $(\vect{p}^{t:t-M},\vect{\bar{p}}^{t+1})$ is encoded by another encoder $E_{\phi_4}$. Their outputs are concatenated and then fed into the encoder $E_{\phi_5}$ to output means and variances of the Gaussian distribution of the latent variable Z. Then we sample Z and concatenate it with the feature of the history condition to generate the input of the decoder $D_{\phi_6}$. Finally, the decoder $D_{\phi_6}$ computes the predicted residual. 

The red connections in \cref{fig:cvae} are only used in the training phase. The ground truth $\vect{r}^{t+1}$ is unavailable during the testing phase. We sample the latent variable Z from a Gaussian distribution $\mathcal{N}(0, \sigma_{latent}\mat{I})$ with a hyper-parameter $\sigma_{latent}$. We extract the history feature from the history condition through the trained encoder $E_{\phi_4}$. Then we concatenate sampled Z and the history feature to decode via the trained decoder $D_{\phi_6}$ to obtain the estimated residual. The neural networks $E_{\phi_3}$, $E_{\phi_4}$, $E_{\phi_5}$, and $D_{\phi_6}$ are all multi-layer perceptrons (MLPs). The dimensions of neural networks and the values of hyper-parameters are shown in \cref{tab:cvae}. 
\begin{table}[tb]
\begin{center}
\caption{\textbf{Details of the CVAE}. The dimensions of every MLP and the values of hyper-parameters in the CVAE are demonstrated.}
\label{tab:cvae}
\small
\resizebox{0.35\textwidth}{!}{
\begin{tabular}{c|c}
\hline
& Dimensions or Values \\
\hline
$E_{\phi_3}$ & [2, 8, 16, 16] \\
\hline
$E_{\phi_4}$ & [18, 512, 256, 16] \\
\hline
$E_{\phi_5}$ & [32, 8, 50, 32] \\
\hline
$D_{\phi_6}$ & [32, 1024, 512, 1024, 2] \\
\hline
M & 7 \\
\hline
$\sigma_{latent}$ & 1.3 \\
\hline
\end{tabular}
}
\end{center}
\end{table}

\subsection{Loss Function and Bayesian Inference}
\label{sec:inference}
Now we are ready to derive the loss function for our BNSP-SFM model to learn the posterior $p(\eta\mid\mat{P})\propto p(\mat{P}\mid\eta)p(\eta)$, where $p_{\phi}(\mat{P}\mid\eta)$ is a doubly stochastic differential equation (\cref{eq:dsde}). Here, $\mat{P}$ denotes the training set and $\eta = \{k_{goal}, k_{col_{ij}}, k_{env} \}$ is directly explainable. In addition, the distributions are parameterized by $\phi = \phi_{\vect{F}} \cup \phi_{\func{\varepsilon}}  = \{\phi_1, \phi_2, \mu_{obj}, \sigma_{obj} \} \cup \{ \phi_3, \phi_4, \phi_5, \phi_6 \}$, which includes unexplainable network weights and distribution parameters that cannot be explained directly. Our learning scheme consists of two parts depending on two losses $\mathcal{L}_{Bayes}$ and $\mathcal{L}_{cvae}$. We use Bayesian inference~\citep{blundell2015weight} to derive $\mathcal{L}_{Bayes}$ because the integral on $p(\eta)$ is intractable to learn $p(\eta\mid\mat{P})$ directly. Specifically, we minimize the KL divergence between a variational posterior $q_{\phi_{\vect{F}}}(\eta)$ and the true posterior $p(\eta\mid\mat{P})$:
\begin{align}
\label{eq:inference}
    &\phi_{\vect{F}} = \argmin_{\phi_{\vect{F}}}D_{\mathbb{KL}}(q_{\phi_{\vect{F}}}(\eta) \| p(\eta\mid\mat{P})) \notag \\
    &= \argmin_{\phi_{\vect{F}}}\mathbb{E}_{q_{\phi_{\vect{F}}}(\eta)} 
    \left[ 
    \log q_{\phi_{\vect{F}}}(\eta) - 
    \log \left( \frac{p(\mat{P}\mid\eta)p(\eta)}{p(\mat{P})}\right) 
    \right] \notag \\
    &= \argmin_{\phi_{\vect{F}}}\{\mathbb{E}_{q_{\phi_{\vect{F}}}(\eta)} [\log q_{\phi_{\vect{F}}}(\eta) -\log p(\mat{P}\mid\eta)p(\eta)] \notag \\
    & \qquad \qquad \quad + \underbrace{\log p(\mat{P})}_{\mbox{const}} \} \notag \\
    &=\argmin_{\phi_{\vect{F}}}\underbrace{\mathbb{E}_{q_{\phi_{\vect{F}}}(\eta)} [\log q_{\phi_{\vect{F}}}(\eta) -\log p(\mat{P}\mid\eta)p(\eta)]}_{\mathcal{L}_{Bayes}(\phi_{\vect{F}}\mid\mat{P},\eta)},
\end{align}
where we assume that both of $q_{\phi_{\vect{F}}}(\eta)$ and $p(\eta)$ are diagonal Gaussian distributions. Then we can get: 
\begin{equation}
\begin{split}
    &\log q_{\phi_{\vect{F}}}(\eta) = \sum_{k \in \eta} -\frac{(k-\mu_{\phi_{\vect{F}}})^2}{2\sigma^2_{\phi_{\vect{F}}}} - \log \sigma_{\phi_{\vect{F}}} - \log \sqrt{2\pi} \\
    &\log p(\eta) = \sum_{k \in \eta} -\frac{(k-\mu_{prior})^2}{2\sigma^2_{prior}} - \log \sigma_{prior} - \log \sqrt{2\pi},
\end{split}
\end{equation}
where $\mu_{\phi_{\vect{F}}}$ and $\sigma^2_{\phi_{\vect{F}}}$ are the means and variances predicted by our method with parameters $\phi_{\vect{F}}$ for each $k \in \eta$. We empirically choose the prior for training. We also model the likelihood $p(\mat{P}\mid\eta)$ as a diagonal Gaussian:
\begin{equation}
\begin{split}
    \log p(\mat{P}\mid\eta) = \sum_{\vect{p}^f \in \mat{P}} -\frac{1}{2} \lVert \vect{\bar{p}}^f-\vect{p}^f \rVert_2^2 - \frac{t_f}{2} \log 2\pi,
\end{split}
\end{equation}
where $\vect{\bar{p}}^f=\{ \vect{\bar{p}}^{t_h+1}, \vect{\bar{p}}^{t_h+2}, \cdots, \vect{\bar{p}}^{t_h+t_f}\}$ is calculated via \cref{eq:bde} given $\eta$ and $\vect{p}^h=\{ \vect{p}^{0}, \vect{p}^{1}, \cdots, \vect{p}^{t_h}\}$, and $\vect{p}^f$ is the ground truth. We optimize $\phi_{\vect{F}}$ using $\mathcal{L}_{Bayes}$. This learns the parameters that are not in the CVAE.

Then we train the CVAE by using:
\begin{equation}
\begin{split}
    \mathcal{L}_{cvae} =& \frac{1}{Nt_f}\sum_{i=1}^N\sum_{t=t_h+1}^{t_h+t_f}\{ \lVert \vect{r}_i^t - \vect{\bar{r}}_i^t\rVert_2^2 \\
  &+ \lambda D_{\mathbb{KL}}(q_{\phi_5}(\vect{z}_i^{t} \mid E_{\phi_3},E_{\phi_4})\|\mathcal{N}(0, \mat{I}))\},
\end{split}
\label{eq:cvae_loss}
\end{equation}
where N is the total number of data samples, $\lambda$ is a tradeoff hyper-parameter, $\vect{r}_i^t=\vect{p}_i^t - \vect{\bar{p}}_i^t$ is the residual and $\vect{\bar{r}}_i^t$ is the predicted residual out of $p_{\phi_6}$. Our overall loss function is $\mathcal{L} = \mathcal{L}_{Bayes}+\mathcal{L}_{cvae}$.

\subsection{Implementation Details}
\label{sec:implementation}
We first pre-train the model without the epistemic uncertainty by using $\mathcal{L}_{Bayes}$ to ensure that the aleatoric uncertainty captures most of the behavior, then train the CVAE via $\mathcal{L}_{cvae}$ while fixing other parameters. Details are shown in \cref{alg:training}.

\begin{algorithm}
\caption{Training strategy}
\begin{algorithmic}
\State \textbf{Parameters}:
\State \qquad Goal Network ($\phi_1$)
\State \qquad Collision Network ($\phi_2$)
\State \qquad Environment Distribution ($\{\mu_{obj}, \sigma_{obj} \}$)
\State \qquad CVAE($\{ \phi_3, \phi_4, \phi_5, \phi_6 \}$)
\State \textbf{Phase 1}: Only Aleatoric Uncertainty
\While{not converged}
\State 1 Train $\phi_1$ via $\mathcal{L}_{Bayes}$ while fixing $\{\phi_2,\mu_{obj}, \sigma_{obj} \}$
\State 2 Train $\{\phi_2,\mu_{obj}, \sigma_{obj} \}$ via $\mathcal{L}_{Bayes}$ while fixing $\phi_1$ 
\EndWhile
\State \textbf{Phase 2}: Add Epistemic Uncertainty
\While{not converged}
\State \quad Train $\{ \phi_3, \phi_4, \phi_5, \phi_6 \}$ via $\mathcal{L}_{cvae}$ 
\State \quad while fixing $\{\phi_1, \phi_2,\mu_{obj}, \sigma_{obj} \}$  
\EndWhile
\end{algorithmic}
\label{alg:training}
\end{algorithm}

We use ADAM for training. The learning rates for the attraction network, the evasion network, and the environment distribution are between $3 \times 10^{-6}$ and $3 \times 10^{-5}$. The learning rates for CVAE are between $1 \times 10^{-7}$ and $1 \times 10^{-6}$.

\begin{table*}[tb]
\begin{center}
\caption{\textbf{Results on SDD based on standard-sampling}. XX/XX is ADE/FDE. BNSP-SFM achieves the state-of-the-art performance in both ADE and FDE compared with all baseline methods. Reported errors are in pixels, and lower is better.}
\label{tab:sdd_standard}
\resizebox{\textwidth}{!}{
\begin{tabular}{c|c|c|c|c|c|c|c|c|c}
\toprule
S-GAN  & Sophie & P2TIRL  & SimAug & PECNet & Y-net  & SocialVAE & V$^{2}$-Net & NSP-SFM  & Ours  \\  
\hline
27.23/41.44 & 16.27/29.38  & 12.58/22.07 & 10.27/19.71 & 9.96/15.88  & 7.85/11.85 & 8.10/11.72 & 7.12/11.39 & 6.52/10.61 & \textbf{6.46}/\textbf{10.49}  \\
\bottomrule
\end{tabular}
}
\end{center}
\end{table*}

\begin{table*}[tb]
\small
\begin{center}
\caption{\textbf{Results on ETH/UCY based on standard-sampling}. XX/XX is ADE/FDE. Our model achieves the state-of-the-art results in ADE and FDE. The unit is meters, and lower is better.}
\label{tab:eth-standard}
\resizebox{\textwidth}{!}{
\begin{tabular}{c|c|c|c|c|c|c|c|c|c}
\toprule
& S-GAN & Sophie & NEXT & PECNet & Y-net & SocialVAE &V$^{2}$-Net & NSP-SFM & BNSP-SFM (Ours)      \\ \midrule
ETH & 0.81/1.52 & 0.70/1.43 & 0.73/1.65 & 0.54/0.87 & 0.28/0.33 & 0.41/0.58  & \textbf{0.23}/0.37 & 0.25/\textbf{0.24} & 0.25/0.25   \\ 
HOTEL & 0.72/1.61 & 0.76/1.67 & 0.30/0.59 & 0.18/0.24 & 0.10/0.14 & 0.13/0.19 & 0.11/0.16 & 0.09/0.13 & \textbf{0.09}/\textbf{0.11}  \\ 
UNIV & 0.60/1.26 & 0.54/1.24 & 0.60/1.27 & 0.35/0.60 & 0.24/0.41 & 0.21/0.36 & 0.21/\textbf{0.35} & 0.21/0.38  & \textbf{0.20}/0.38    \\ 
ZARA1 & 0.34/0.69 & 0.30/0.63 & 0.38/0.81 & 0.22/0.39 & 0.17/0.27 & 0.17/0.29 & 0.19/0.30 & 0.16/0.27 & \textbf{0.16}/\textbf{0.27} \\ 
ZARA2 & 0.42/0.84 & 0.38/0.78 & 0.31/0.68 & 0.17/0.30 & 0.13/0.22 & 0.13/0.22 & 0.14/0.24 & 0.12/0.20  & \textbf{0.12}/\textbf{0.19} \\ \hline \rule{0pt}{\normalbaselineskip}AVG & 0.58/1.18 & 0.54/1.15 & 0.46/1.00 & 0.29/0.48 & 0.18/0.27 & 0.21/0.33 &  0.18/0.28  & 0.17/0.24 & \textbf{0.16}/\textbf{0.24}\\ \bottomrule
\end{tabular}
}
\end{center}
\end{table*}
\section{Experiments}
\subsection{Datasets}
We use two public datasets for evaluation: SDD~\citep{robicquet2016learning} and ETH/UCY~\citep{pellegrini2010improving,lerner2007crowds}, which are widely used in human trajectory forecasting. \textbf{Stanford Drone Dataset (SDD):} The dataset contains videos across 20 different scenes in bird’s eye view, with more than 100,000 pedestrian-pedestrian interactions and pedestrian-environment interactions. Following \citep{mangalam2021goals}, we extract trajectories with a time step 0.4 seconds and obtain 20-frame samples for an 8/12 setting, i.e., given the first 8 frames (3.2 seconds, $t_h=7$), we aim to predict the future 12 frame trajectories (4.8 seconds, $t_f=12$). \textbf{ETH/UCY Datasets:} There are five sub-datasets (ETH, Hotel, Univ, Zara1, and Zara2), including more than 1500 pedestrians with various behaviors such as collision avoidance. Following the standard leave-one-out evaluation protocol~\citep{gupta2018social,mangalam2020not,mangalam2021goals}, we train our model on four sub-datasets and test it on the remaining one in turn. The world coordinates used by the dataset don't match some parts of our model such as the goal sampling network and $\func{\mathcal{F}}_{env}$. Our model generally works in the pixel space. Therefore, we convert the world coordinates into pixel coordinates using the homography matrices from Y-net~\citep{mangalam2021goals}. We project the predictions back into the world space to calculate errors for fair comparisons with existing methods. We extract the trajectories in the same way as SDD and adopt the same 8/12 prediction strategy.  For pedestrians that have less than 20 frames, we treat them as observed dynamic obstacles as part of the environment.

\subsection{Trajectory Forecasting}

We adopt well-established Average Displacement Error (ADE) and Final Displacement Error (FDE)~\citep{alahi2016social,gupta2018social,Yue_trajectory_2022} to measure the prediction accuracy. ADE is the $\ell_2$ error between a predicted trajectory and its ground truth averaged over all positions, while FDE is the $\ell_2$ error between the predicted destination and its ground truth. Following prior research, we report the best ADE and FDE among multiple sampled trajectories. There are two existing strategies in existing work: standard-sampling and ultra-sampling. Standard-sampling employs 20 sampled trajectories and ultra-sampling employs 20 sampled positions at each step. The former is more widely employed and the latter is employed when the model is intrinsically stochastic especially when multiple stochastic components exist~\citep{Yue_trajectory_2022}. We compare our BNSP-SFM in both standard-sampling and ultra-sampling with a wide range of baselines: Social GAN (S-GAN)~\citep{gupta2018social}, Sophie~\citep{sadeghian2019sophie}, NEXT~\citep{liang2019peeking}, P2TIRL~\citep{deo2020trajectory}, SimAug~\citep{liang2020simaug},  PECNet~\citep{mangalam2020not}, Y-Net~\citep{mangalam2021goals}, S-CSR~\citep{zhou2021sliding}, SocialVAE~\citep{xu2022socialvae}, V$^2$-Net~\citep{wong2022view}, and NSP-SFM~\citep{Yue_trajectory_2022}. 

We first show the standard-sampling results in \cref{tab:sdd_standard} and \cref{tab:eth-standard}. Our BNSP-SFM achieves state-of-the-art performance on both SDD and ETH/UCY. Overall, both NSP-SFM and BNSP-SFM outperform other methods. BNSP-SFM provides slightly better results than NSP-SFM. On SDD, BNSP-SFM obtains 6.46/10.49 in ADE/FDE, improving the past state-of-the-art model NSP-SFM (6.52/10.61) by 0.92\%/1.13\%. Compared with other previous methods, our BNSP-SFM has at least 9.27\%/7.90\% improvement in ADE/FDE. On ETH/UCY, the improvement is 5.88\% in ADE on average and gains a better average FDE with the maximum 15.38\% improvement in Hotel. Since BNSP-SFM is based on NSP-SFM, it is understandable that they can achieve similar numerical accuracy. However, the Bayesian components do not hurt prediction accuracy while providing additional explainability and confidence estimation, which are extra benefits.

Further, we show ultra-sampling results in \cref{tab:ultra}. BNSP-SFM outperforms NSP-SFM by approximately 14.61\%/60.17\% in ADE/FDE on SDD. We also observe that BNSP-SFM gains the same performance in ADE and 50\% improvement in FDE on ETH/UCY compared with NSP-SFM. BNSP-SFM has higher performance than S-SCR on both datasets. Stochastic models with ultra-sampling such as S-CSR and NSP-SFM have shown to be more accurate in prediction~\citep{Yue_trajectory_2022}, but BNSP-SFM still outperform them in general. Further, for a fair comparison with S-CSR, both BNSP-SFM and NSP-SFM sample 20 destinations and only sample 15 positions at each step, so the total number of sampled trajectories is slightly smaller than S-CSR. S-CSR doesn't need the given goals and samples 20 positions at each step following the ultra-sampling strategy. However, both BNSP-SFM and NSP-SFM outperform S-CSR. The main difference is that S-CSR is a black-box neural network based on Variational Autoencoder, while the core of BNSP-SFM and NSP-SFM are based on explicit models, which clearly demonstrates its advantages. Further, BNSP-SFM improves the accuracy or is at least in par with NSP-SFM in all scenarios, indicating that the Bayesian components manage to capture the uncertainty better and our new model retains the great prediction ability. 
\begin{table}[tb]
\small
\begin{center}
\caption{\textbf{Results on ETH/UCY and SDD based on ultra-sampling}. XX/XX is ADE/FDE. Our model BNSP-SFM outperforms other baselines on all datasets in ADE and FDE.}
\label{tab:ultra}
\resizebox{0.475\textwidth}{!}{
\begin{tabular}{c|c|c|c}
\toprule
     &   S-CSR  & NSP-SFM & BNSP-SFM (Ours)  \\  \hline\rule{0pt}{\normalbaselineskip}
    ETH  & 0.19/0.35 & 0.07/0.09 & \textbf{0.05}/\textbf{0.04}  \\
     HOTEL & 0.06/0.07 & 0.03/0.07 & \textbf{0.03}/\textbf{0.02} \\
     UNIV & 0.13/0.21 & 0.03/0.04 & \textbf{0.03}/\textbf{0.03}\\
     ZARA1 & 0.06/0.07 & 0.02/0.04 & \textbf{0.02}/\textbf{0.02}\\
     ZARA2 & 0.06/0.08 & 0.02/0.04 & \textbf{0.02}/\textbf{0.02}\\
     \hline 
     AVG & 0.10/0.16 & 0.03/0.06 &\textbf{0.03}/\textbf{0.03}\\
     \hline \hline
     SDD & 2.77/3.45 & 1.78/3.44 & \textbf{1.52}/\textbf{1.37}\\ \bottomrule
\end{tabular}
}
\end{center}
\end{table}


\subsection{Generalization}
\subsubsection{Generalization on Cross-scene Testing}
\begin{table}[tb]
\begin{center}
\caption{\textbf{Results across scenarios}. All models are trained on ETH/UCY and tested on SDD. Our model achieves the best performance. Reported errors are in pixels, and lower is better.}
\label{tab:across}
\resizebox{0.48\textwidth}{!}{
\begin{tabular}{c|c|c|c}
\toprule
     Models & Y-net  & NSP-SFM & BNSP-SFM (Ours)  \\  \hline\rule{0pt}{\normalbaselineskip}
    ADE/FDE & 30.59/51.43 & 6.65/10.60 & \textbf{6.55}/\textbf{10.59}  \\
     \bottomrule
\end{tabular}}
\end{center}
\end{table}
One way to test the generalizability is cross-scene testing, \ie, training on one scene and testing on another, drastically different scene. Although the results on ETH/UCY and SDD are already based on cross-scene testing, we increase the challenge by using ETH/UCY and SDD as training and testing data respectively, where the scenes are more different in terms of the environment, space size and pedestrian dynamics. For comparison, we choose Y-net and NSP-SFM as baselines and show the results in \cref{tab:across}. The performance of Y-net drops severely from 7.85/11.85 (when it's trained also on SDD) to 30.59/51.43. NSP-SFM and BNSP-SFM perform slightly worse than when they are also trained on SDD, but considerably better than Y-net. NSP-SFM changes from 6.52/10.61 to 6.65/10.60 while BNSP-SFM changes from 6.46/10.49 to 6.55/10.59. This shows NSP-SFM and BNSP-SFM learn intrinsic behaviors of pedestrians that are universal across scenes.

\subsubsection{Generalization to High-density Scenarios}
Generalization is the ability to adapt to unseen data. Normally, we assume that the distribution of the training data is similar to that of the testing data. A model with a good generalizability should perform well not only on the testing data but also on drastically different data. For trajectory prediction methods, an efficient way to evaluate their generalizability is to use the trained model to simulate pedestrians with much higher densities, as it will lead to significantly different pedestrian dynamics~\citep{narang2015generating}. 

Following~\citep{Yue_trajectory_2022}, we adopt the collision rate as the metric to evaluate the generalizability. We regard each agent as a disc with a radius of $r$ pixels. We count one collision if the minimum distance between two trajectories falls below 2$r$ at any time. Given N agents in the scene, we calculate the collision rate as $R_{col} = \frac{M}{N(N-1)/2}$, where M is the number of collisions. Generally, the ground-truth $r$ is hard to acquire because of the tracking error, the distorted images, etc. We observe unrealistically high collision rates in all cases when $r$ is too large. In contrast, the collision rate will always be too low when $r$ is too small, \eg, $r = 0$ will give 0\% collision rate all the time. Therefore, we need to search for a reasonable $r$ that keeps the collision rate of the ground-truth data approximately zero. 

According to ~\citep{Yue_trajectory_2022}, we set the $r$ as $7.5$ pixels $0.2 \; m$ on SDD and ETH/UCY, respectively. A lower collision rate means higher plausibility and better generalization. We conduct two kinds of experiments: one is the collision rates on the testing data of SDD and ETH/UCY; the other is the collision rates on unseen scenarios with higher crowd densities. The first one compares the plausibility of predicted trajectories in scenarios similar to the training data. The second one pushes the models for generalization. We use NSP-SFM, Y-net, and S-CSR as baselines.
\begin{table}[tb]
\begin{center}
\caption{\textbf{Collision rates on the
testing data of SDD and ETH/UCY}. BNSP-SFM achieves the same or better performance on all datasets.}
\label{tab:collision_test}
\small
\resizebox{0.475\textwidth}{!}{
\begin{tabular}{c|c|c|c|c}
\toprule
     & Y-net &  S-CSR  & NSP-SFM & BNSP-SFM (Ours)  \\  \hline\rule{0pt}{\normalbaselineskip}
    ETH & 0 & 0 & 0 & \textbf{0}  \\
     HOTEL & 0 & 0 & 0 & \textbf{0}\\
     UNIV & 1.51\% & 1.82\% & 1.48\% & \textbf{1.48\%}\\
     ZARA1 & 0.82\% & 0.41\% & 0 & \textbf{0}\\
     ZARA2 & 1.31\% & 1.31\% & 0.66\% & \textbf{0.66\%} \\
     \hline 
     AVG & 0.73\% & 0.71\% & 0.43\% & \textbf{0.43\% }\\
     \hline \hline
     SDD & 0.47\% & 0.42\% & 0.42\% & \textbf{0.40\%} \\ \bottomrule
\end{tabular}
}
\end{center}
\end{table}

Collision rates on testing data (\cref{tab:collision_test}) show all methods achieve reasonable results, with the maximum 1.82\% from S-CSR on UNIV. All methods achieve 0\% in ETH and HOTEL. This is not surprising for two reasons. First, all methods predict well. So, if the ground-truth data does not contain collisions, neither do the predictions. The second reason is the sparsity of people in the scene. For instance, the highest number of people who are simultaneously in the scene is 11 in Coupa0 (in SDD). Even if the prediction goes wrong, there might not be people around, so no collisions will occur. When people are indeed close to each other, NSP-SFM and BNSP-SFM outperform Y-net and S-CSR. This is because our learned explicit model has an explicit repulsive force between agents, and can therefore avoid collisions.

\begin{figure}[tb]
\centering
\includegraphics[width=0.9\linewidth]{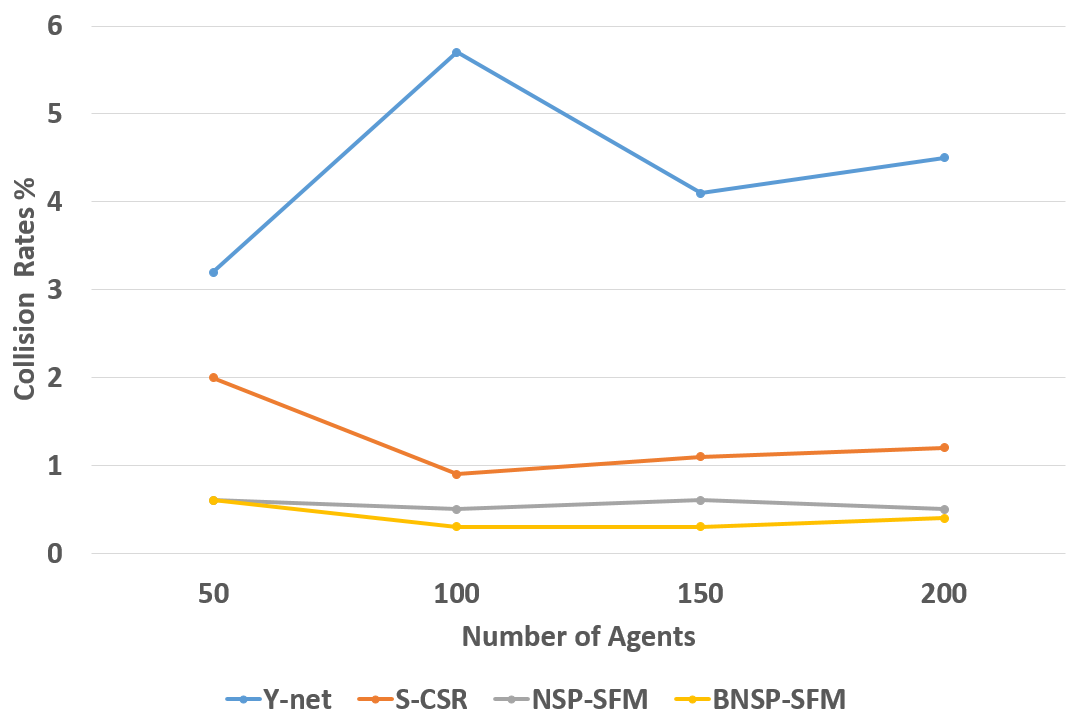}
\caption{\textbf{The relationship between the collision rate and the number of agents}. Our BNSP-SFM achieves the lowest collision rates across all settings.}
\label{fig:col_rate}
\end{figure}  
\begin{figure}[tb]
\begin{center}
\includegraphics[width = \linewidth]{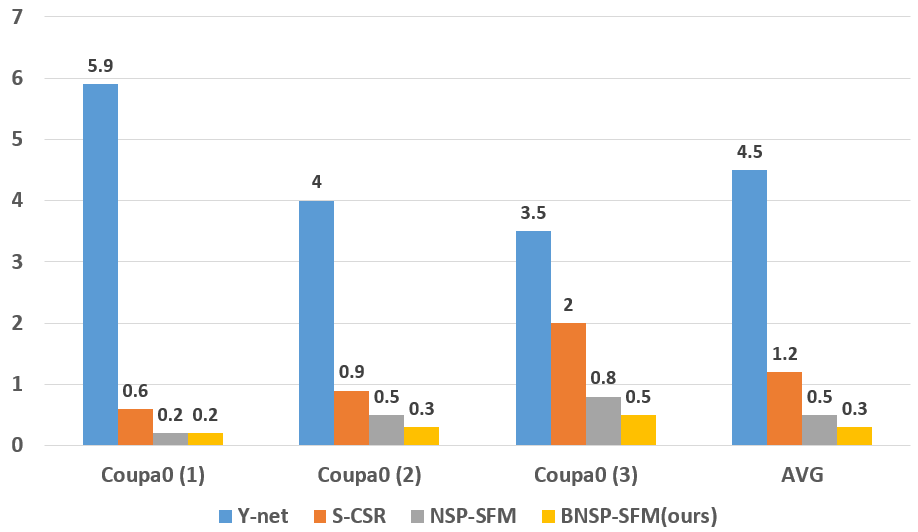}
\end{center}
\caption{\textbf{Collision rates on Coupa0 with 200 HNP}. Three collision rates on three intervals and their average are reported for all methods. Our BNSP-SFM outperforms all baseline methods and shows the stronger generalizability. The unit for the vertical axis is percentage.}
\label{fig:collision_coupa0}
\end{figure}

To test models in drastically different scenarios from the two datasets, we select the scene Coupa0 from SDD and use the highest number of people (HNP) in the scene as an indicator of the crowd density. Coupa0 has a large space so, in theory, it can contain many people. However, the HNP in the original Coupa0 is merely 11. Therefore, we increase the HNP to 50, 100, 150, and 200 people. Then we use all methods trained on the original SDD data as simulators to run a long simulation and compute the collision rates for three time intervals of the simulation under each HNP, following the same setting in~\citep{Yue_trajectory_2022}. Specifically, instead of creating all agents at the same time, we initiate agents batch by batch from the boundaries so that they start with no collisions and walk into the scene. For 30-second simulations, we divide the time into t = 0 to 8, t = 4 to 12, and t = 8 to 16, where the density in the central area is the highest during t=8 to 16.

We report the average collision rates of three intervals in each simulation for every method in \cref{fig:col_rate}. Overall, BNSP-SFM and NSP-SFM outperform the baseline methods with lower collision rates across different agent numbers. To further understand it, we show detailed collision rates on three intervals for all methods in \cref{fig:collision_coupa0} when HNP=200. Y-net performs poorly in all three time intervals. This indicates that, from the very beginning, collisions start to happen. Comparatively, S-CSR performs better, especially in the beginning when the agents start to walk into the scene. However, its collision rate spikes to 2\% when the density becomes the highest. Comparatively, BNSP-SFM and NSP-SFM perform well in every time interval. Although their collision rates also increase, the highest is only 0.8\% for NSP-SFM and 0.5\% for BNSP-SFM. On average, Y-net, S-CSR, and NSP-SFM are 1400\%, 300\%, and 66.67\% worse than BNSP-SFM. 

\begin{figure}[tb]
\centering
\includegraphics[width=0.9\linewidth]{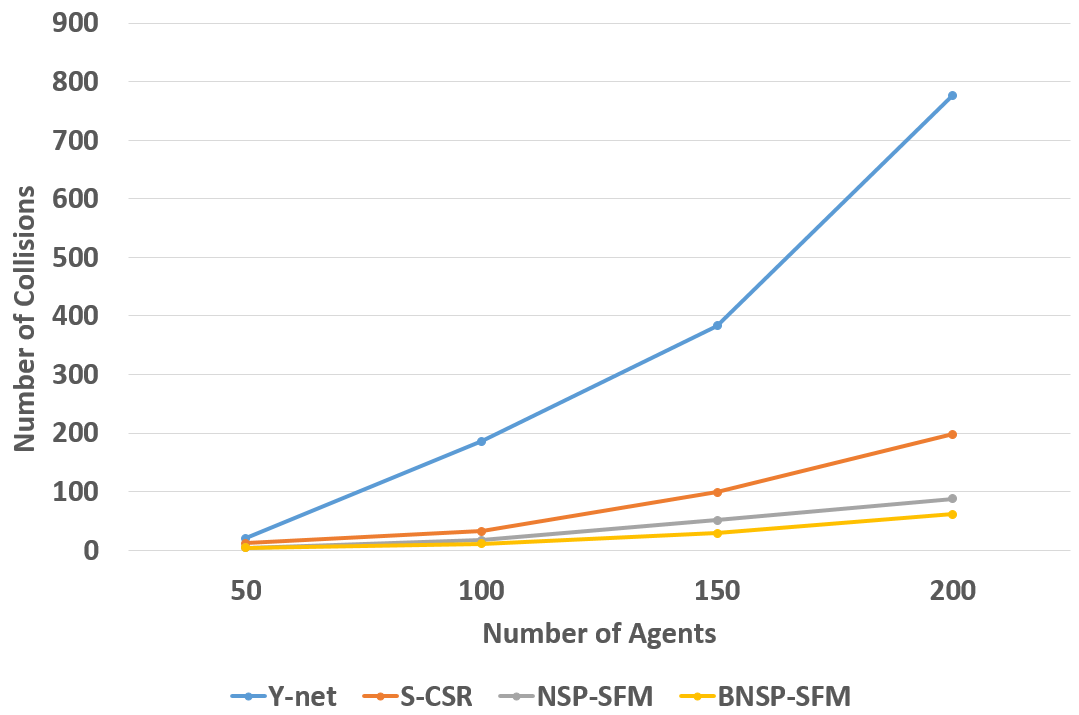}
\caption{\textbf{The relationship between the number of collisions and the number of agents}. Our BNSP-SFM outperforms other baseline methods in the number of collisions.}
\label{fig:col_number}
\end{figure}

Collision rate does not directly reflect the number of collisions. In safety-critical applications, every collision should count. Therefore, we also show the averaged number of collisions of three intervals in each simulation experiment in \cref{fig:col_number}. All models have more collisions when the density becomes higher. However, our BNSP-SFM model maintains the lowest collision number across all simulation settings. The closest second is NSP-SFM. Moreover, BNSP-SFM also shows the lowest increase rate of collisions when the density increases. Overall, we demonstrate that BNSP-SFM and NSP-SFM possess the strongest generalization to high-density scenarios through collision rates and the number of collisions. The repulsive force in our explicit model plays a key role in collision avoidance.

\subsection{Explainability of Prediction}
\begin{figure}[tb]
\begin{center}
\includegraphics[width =\linewidth]{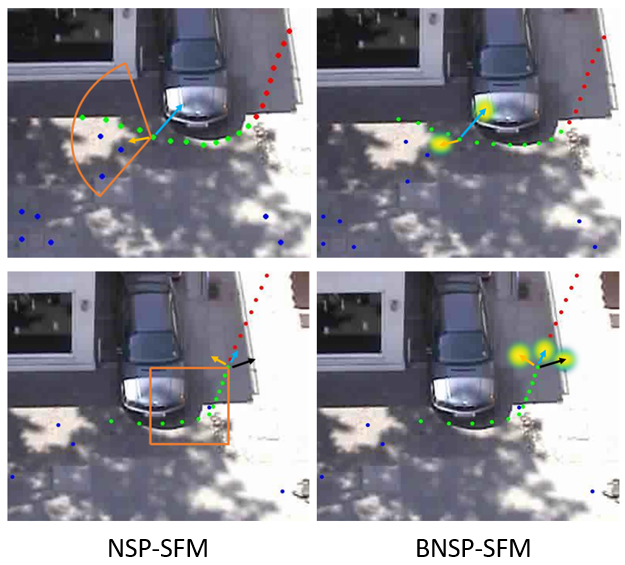}
\end{center}
\caption{\textbf{Explainability of Behaviors}. Red, blue, and green dots denote observations, neighbors, and predictions respectively. Main factors $\func{\mathcal{F}}_{goal}$, $\func{\mathcal{F}}_{col}$, and $\func{\mathcal{F}}_{env}$ are shown as yellow, light blue, and black arrows for a pedestrian. The orange areas are two view fields in~\citep{Yue_trajectory_2022}. The major difference between BNSP-SFM and NSP-SFM is the confidence in the factors shown as heatmaps.}
\label{fig:inter}
\end{figure}
Being able to explain behaviors is crucial in human trajectory forecasting~\citep{wang_trending_2016,Helbing_Social_1995}. Our BNSP-SFM not only provides a plausible interpretation for pedestrian behaviors, but also gives the confidence of the interpretation. To this end, we analyze the three main factors $\func{\mathcal{F}}_{goal}$, $\func{\mathcal{F}}_{col}$, $\func{\mathcal{F}}_{env}$ in the behavioral model, all of which are Gaussians. We demonstrate several explainability examples and compare this model with NPS-SFM~\citep{Yue_trajectory_2022}. 

In~\cref{fig:inter}, we choose pedestrians with similar trajectories predicted by BNSP-SFM and NSP-SFM. In the top row, the pedestrian steers to avoid other pedestrians instead of going directly towards the goal. Both methods explain the steering by the influence of the goal attraction and the collision avoidance (yellow and blue arrows). However, BNSP-SFM also gives the confidence of the explanation, shown as heatmaps based on the learned means and variances of different factors. This not only gives the possible alternative explanations, but also shows how confident BNSP-SFM is regarding each explanation. 

\cref{fig:inter} Bottom shows a slightly more complex explanation where there is also environment repulsion (black arrow). The agent is attracted by his/her destination (yellow arrow) while he/she avoids collisions with other pedestrians (light blue arrow) and is repelled by the obstacle car in the environment (black arrow). When there are multiple factors involved, the confidence maps associated with factors are more informative in term of their relative confidence in our model. The environment repulsion arises when the person is very close to the car and avoiding the collision becomes a major concern. This is reflected by the environment repulsion having a more concentrated confidence map than the goal attraction and the collision avoidance. To be specific, the standard deviations for the factors  $\func{\mathcal{F}}_{goal}$, $\func{\mathcal{F}}_{col}$, and $\func{\mathcal{F}}_{env}$ here are $[9.13, 11.40], [5.73, 8.14], [5.29, 2.02]$, respectively. Two standard deviations for each factor correspond to x-axis and y-axis. This means BNSP-SFM is more certain about the influence of the environment repulsion. 

Overall, both NSP-SFM and BNSP-SFM give similar predicted trajectories with great prediction accuracy. Both methods can explain the same three factors. However, the means and the variances of the estimated distributions of social forces in BSNP-SFM provide more informative and explainable predictions.   

\begin{figure}[tb]
\centering
\includegraphics[width=0.8\linewidth]{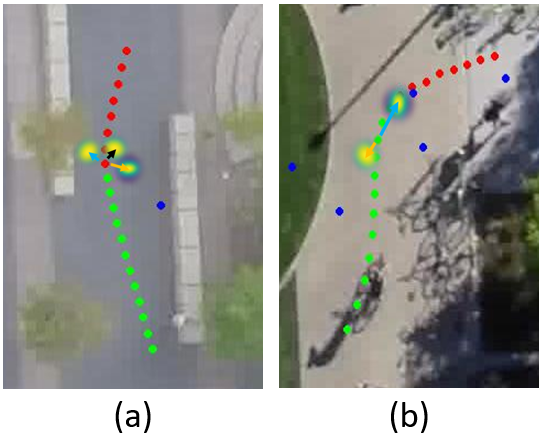}
\caption{\textbf{Explainability of Behaviors}. Red, blue, and green dots denote observations, neighbors, and predictions, respectively. Main factors $\func{\mathcal{F}}_{goal}$, $\func{\mathcal{F}}_{col}$, $\func{\mathcal{F}}_{env}$ are shown as yellow, light blue, and black arrows for a pedestrian. The confidence in the factors is shown as heatmaps.}
\label{fig:inter_behavior}
\end{figure} 

We show more explainability examples of our BNSDE model in~\cref{fig:inter_behavior}. The future trajectories (green dots) are predicted by using the standard sampling, where the $\func{\mathcal{F}}_{goal}$ dominates among the three factors and has a more concentrated confidence map, shown in \cref{fig:inter_behavior} (a). This is likely because there are not many imminent collisions for this person. With the more concentrated confidence map, BNSP-SFM is more certain about the attraction of the destination, and the predicted future trajectory goes almost straight to the destination. In \cref{fig:inter_behavior} (b), we ignore the influence of environment $\func{\mathcal{F}}_{env}$ because it's too weak to be visualized here. $\func{\mathcal{F}}_{col}$ dominates and has a more concentrated confidence map, meaning that our BNSP-SFM model is more certain about collision avoidance. This is because the neighbor in front of the person is walking at a high speed. The predicted future trajectory first avoids the neighbor then aims for the destination.  

In human trajectory forecasting, to our best knowledge, few deep learning methods~\citep{hossain22sfmgnet,kreiss2021deep,Yue_trajectory_2022} can provide explainability. In theory, we can also visualize black-box deep learning methods such as latent features or layer activations. However, it's difficult to determine how to visualize them to explain the behaviors. Among the explainable models, BNSP-SFM models the fine-grained structure of commonly observed uncertainty~\citep{He_Informative_2020} and is therefore more explainable.
\begin{figure}[tb]
\centering
\includegraphics[width=0.8\linewidth]{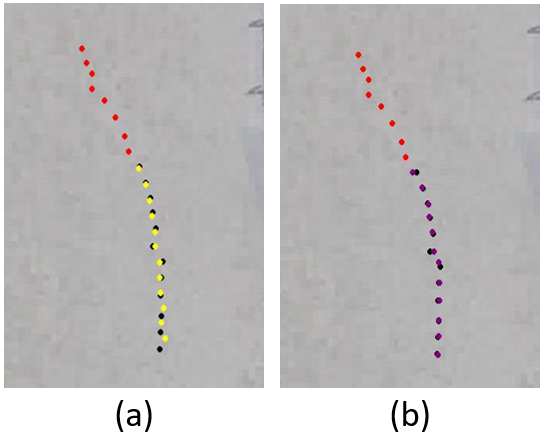}
\caption{\textbf{Aleatoric and epistemic uncertainty}. Red and black dots denote observations and the ground truth future trajectory. Yellow dots denote the prediction with only aleatoric uncertainty. Purple dots denote the prediction with aleatoric and epistemic uncertainty.}
\label{fig:inter_uncertainty}
\end{figure} 

\subsection{Epistemic Uncertainty}
\cref{fig:inter_uncertainty} shows how the BNSP-SFM model captures epistemic uncertainty as a residual of the predicted behavior with the aleatoric uncertainty. There are no neighbors or obstacles. The predicted trajectories in \cref{fig:inter_uncertainty} (yellow and purple dots) are based on ultra-sampling. We capture only aleatoric uncertainty in \cref{fig:inter_uncertainty} (a). We can see that the yellow dots are close to the ground truth (black dots), showing partial uncertainty has been captured. BNSP-SFM exploits CVAE to capture the remaining epistemic uncertainty. Purple dots in \cref{fig:inter_uncertainty} (b) denote the prediction with both the aleatoric and the epistemic uncertainty and almost overlap with the ground truth. This means that our model can capture the complete uncertainty well.

\subsection{Data Efficiency}
For human trajectory forecasting, it is expensive and time-consuming to collect clean data, which might involve manual labelling and checking. Therefore, data efficiency is crucial. We conduct experiments to test the data efficiency. We decrease the training data of SDD to 50\% and 25\%, then train BNSP-SFM, NSP-SFM, and S-CSR and test them on the original testing data of SDD. The results are shown in \cref{tab:data-efficiency}. 

Both BNSP-SFM and NSP-SFM have higher data efficiency than S-CSR. Intuitively, this is sensible. When there is little data, black-box deep neural networks like S-CSR will be under-trained or overfitted. Having an explicit model in neural nets significantly reduces the required data size, which is known in differentiable physics models as explicit models can act as regularizers in learning~\citep{Gong_finegrained_2022}. Both NSP-SFM and BNSP-SFM benefit from it.  

Between NSP-SFM and BNSP-SFM, BNSP-SFM has less performance deterioration when the training data is reduced. This is because BNSP-SFM learns distributions of social forces, while NSP-SFM learns deterministic forces. Once the distributions are learned, it is still possible to sample good predictions, while NSP-SFM needs enough data to accurately learn the forces across space and time. 


\begin{table}[tb]
\begin{center}
\caption{\textbf{Results on partial training data of SDD based on ultra-sampling}. XX/XX is ADE/FDE. All methods are trained using partial training data of SDD (from 25\% to 100\%) and are tested on the original testing data of SDD.}
\label{tab:data-efficiency}
\resizebox{0.475\textwidth}{!}{
\begin{tabular}{c|c|c|c}
\toprule
     & S-CSR  & NSP-SFM & BNSP-SFM (Ours)  \\  \hline\rule{0pt}{\normalbaselineskip}
   Full SDD & 2.77/3.45 & 1.78/3.44 & \textbf{1.52}/\textbf{1.37}  \\
   50 \% SDD & 3.78/5.09 & 2.03/4.15 & \textbf{1.89}/\textbf{1.80}  \\
      25 \% SDD & 4.60/6.44 & 2.10/4.32 & \textbf{2.01}/\textbf{2.00}  \\
     \bottomrule
\end{tabular}
}
\end{center}
\end{table}

\begin{figure}[tb]
\centering
\includegraphics[width=1\linewidth]{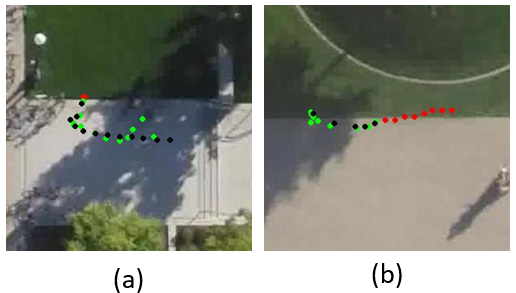}
\caption{\textbf{Failure cases from the BNSP-SFM model}. Red, green, and black dots denote the observations, predictions, and the ground truth future trajectories. (a) The goal prediction has high error leading to an overall low-quality prediction. (b) the pedestrian suddenly stopped resulting in zero velocity for a number frames. Such behaviors are not captured very well as the goal attraction drags the pedestrian to move around the goal before it completely stop there.} 
\label{fig:failure}
\end{figure}

\subsection{Analysis of Failure Cases}
Despite achieving the state-of-the-art performance in multiple applications, there are two sources of prediction error that BNSP-SFM can struggle to mitigate. Similar to deterministic SFM models, our model replies on the destination, which, if predicted wrong, can cause high overall prediction error, shown in \cref{fig:failure} (a). This is decided by the overall explicit nature of BNSP-SFM, i.e. assuming the overall movement trend is largely decided by the force leading to the destination. A more fine-grained modeling of the destination could improve the results. Also, BNSP-SFM cannot predict well on highly nonlinear motions which is rare in the data employed. This is shown in \cref{fig:failure} (b) where the pedestrian suddenly stopped, giving near-zero velocities in a number of frames. In this case, the goal attraction force can cause the agent to overshoot in reaching the goal. As a result, the agent moves around the goal for some time before being able to completely stop there. Such sudden stop behaviors also cause issues in simulators where some heuristics are needed to make the agent to stop as soon as they reach their goals. We do not employ similar heuristics as we mainly target prediction in this paper.


\subsection{Ablation Study}
\begin{table}[tb]
\begin{center}
\caption{\textbf{Ablation study on SDD}. XX/XX is ADE/FDE.}
\label{tab:ablation}
\resizebox{0.475\textwidth}{!}{
\begin{tabular}{cccccc}
\toprule
$\func{\mathcal{F}}_{att}$ & \faCheck  & \faCheck & \faCheck & \faCheck        \\
$\func{\mathcal{F}}_{eva}$ & \faTimes  & \faCheck  & \faCheck & \faCheck     \\
$\func{\mathcal{F}}_{env}$ & \faTimes  & \faCheck  & \faCheck & \faCheck \\
Aleatoric    & \faTimes  & \faTimes & \faCheck  & \faCheck  \\
Epistemic    & \faTimes  & \faTimes  & \faTimes & \faCheck \\
\hline
BNSP-SFM & 6.52/10.57  & 6.46/10.49 & 2.32/2.66 & 1.52/1.37\\
NSP-SFM & 6.57/10.68  & 6.52/10.61 & n/a & 1.78/3.44\\
\bottomrule
\end{tabular}
}
\end{center}
\end{table}

To further understand the different components in our model, we conduct ablation experiments on SDD, shown in \cref{tab:ablation}. We can see that our model can obtain good results when only considering $\func{\mathcal{F}}_{goal}$ because this is the goal attraction, which already determines the motion trend and governs the motion dynamics. $\func{\mathcal{F}}_{col}$ and $\func{\mathcal{F}}_{env}$ might not exist if there is no imminent collision with other pedestrians or the environment. However, we still obtain more accurate results by considering them. 

Next, significant improvement is obtained by incorporating the aleatoric uncertainty. This demonstrates that our aleatoric model can capture the dynamics stochasticity well, meaning the aleatoric uncertainty is also a large source of prediction error. Note that the aleatoric uncertainty is modeled at the force level which further models the social interactions. Capturing such interaction uncertainty significantly improves the prediction.

Finally, we gain an even better result 1.52/1.37 by combining aleatoric and epistemic uncertainty. The total uncertainty is captured well by exploiting the excellent data-fitting capacity of deep learning. We can see that the performance improvement trend is similar for both BNSP-SFM and NSP-SFM in \cref{tab:ablation}. This is because these two methods model the same three factors; the major difference is where the uncertainty is captured. To show the importance of fine-grained modeling of uncertainty, we compare BNSP-SFM with NSP-SFM. One noticeable result is that NSP-SFM also improves when uncertainty is captured, jumping from 6.52/10.61 to 1.78/3.44. However, such a gain is obtained by blindly fitting a neural network to capture the combined aleatoric and epistemic uncertainty. Consequently, not only is the uncertainty unexplainable, the overall performance is also slightly worse.

\section{Discussion, Conclusion and Future Work}
We have proposed a novel Bayesian neural stochastic differentiable equation model for human trajectory forecasting. BNSP-SFM outperforms existing methods, achieving higher prediction accuracy, better generalizability, more explainability and higher data efficiency. One limitation is that our model does not explicitly consider high-level factors such as affective states in crowd dynamics, as studied in crowd research, which has found be to closely relevant to the motion randomness. In future, we will incorporate high-level factors, including the affective state, for better explainability. Moreover, we will explore our model in other areas such as autonomous vehicles and social robots.

\section{Acknowledgment}
This project has received funding from the European Union’s Horizon 2020 research and innovation programme under grant agreement No 899739 CrowdDNA.

\section*{Declarations}
\begin{itemize}
\item Funding. This project has received funding from the European Union’s Horizon 2020 research and innovation programme under grant agreement No 899739 CrowdDNA.
\item Conflict of interest/Competing interests (check journal-specific guidelines for which heading to use). Not applicable
\item Ethics approval. Not applicable
\item Consent to participate. Not applicable
\item Consent for publication. All authors have given their consent for publication.
\item Availability of data and materials. Data sharing not applicable to this article as no datasets were generated during the current study.
\item Code availability. The code will be shared upon acceptance. Part of the code has been shared at http://drhewang.com/pages/NSP.html
\item Authors' contributions. Jiangbei Yue contributed in conceptualization, experiment design, conducting experiments and drafting the paper. Dinesh Manocha contributed in conceptualization and paper drafting. He Wang led the whole research project, including conceptualizing the research idea, supervision, drafting the paper, etc. 
\end{itemize}


\bibliography{sn-bibliography}

\begin{thebibliography}{}
\providecommand{\doi}[1]{\url{https://doi.org/#1}}
\bibcommenthead

\bibitem [\protect \citeauthoryear {%
Alahi%
\ \protect \BOthers {.}}{%
Alahi%
\ \protect \BOthers {.}}{%
{\protect \APACyear {2016}}%
}]{%
alahi2016social}
\APACinsertmetastar {%
alahi2016social}%
\begin{APACrefauthors}%
Alahi, A.%
, Goel, K.%
, Ramanathan, V.%
, Robicquet, A.%
, Fei-Fei, L.%
\BCBL {} Savarese, S.%
\end{APACrefauthors}%
\unskip\
\newblock
\APACrefYearMonthDay{2016}{}{}.
\newblock
{\BBOQ}\APACrefatitle {Social lstm: Human trajectory prediction in crowded
  spaces} {Social lstm: Human trajectory prediction in crowded spaces}.{\BBCQ}
\newblock
 \APACrefbtitle {Proceedings of the IEEE conference on computer vision and
  pattern recognition} {Proceedings of the ieee conference on computer vision
  and pattern recognition}\ (\BPGS\ 961--971).
\PrintBackRefs{\CurrentBib}

\bibitem [\protect \citeauthoryear {%
Bartoli%
, Lisanti%
, Ballan%
\BCBL {}\ \BBA {} Del~Bimbo%
}{%
Bartoli%
\ \protect \BOthers {.}}{%
{\protect \APACyear {2018}}%
}]{%
bartoli2018context}
\APACinsertmetastar {%
bartoli2018context}%
\begin{APACrefauthors}%
Bartoli, F.%
, Lisanti, G.%
, Ballan, L.%
\BCBL {} Del~Bimbo, A.%
\end{APACrefauthors}%
\unskip\
\newblock
\APACrefYearMonthDay{2018}{}{}.
\newblock
{\BBOQ}\APACrefatitle {Context-aware trajectory prediction} {Context-aware
  trajectory prediction}.{\BBCQ}
\newblock
 \APACrefbtitle {2018 24th International Conference on Pattern Recognition
  (ICPR)} {2018 24th international conference on pattern recognition (icpr)}\
  (\BPGS\ 1941--1946).
\PrintBackRefs{\CurrentBib}

\bibitem [\protect \citeauthoryear {%
Bendali-Braham%
, Weber%
, Forestier%
, Idoumghar%
\BCBL {}\ \BBA {} Muller%
}{%
Bendali-Braham%
\ \protect \BOthers {.}}{%
{\protect \APACyear {2021}}%
}]{%
bendali2021recent}
\APACinsertmetastar {%
bendali2021recent}%
\begin{APACrefauthors}%
Bendali-Braham, M.%
, Weber, J.%
, Forestier, G.%
, Idoumghar, L.%
\BCBL {} Muller, P\BHBI A.%
\end{APACrefauthors}%
\unskip\
\newblock
\APACrefYearMonthDay{2021}{}{}.
\newblock
{\BBOQ}\APACrefatitle {Recent trends in crowd analysis: A review} {Recent
  trends in crowd analysis: A review}.{\BBCQ}
\newblock
\APACjournalVolNumPages{Machine Learning with Applications}{4}{}{100023}.
\newblock

\newblock

\PrintBackRefs{\CurrentBib}

\bibitem [\protect \citeauthoryear {%
Bennewitz%
, Burgard%
\BCBL {}\ \BBA {} Thrun%
}{%
Bennewitz%
\ \protect \BOthers {.}}{%
{\protect \APACyear {2002}}%
}]{%
bennewitz2002learning}
\APACinsertmetastar {%
bennewitz2002learning}%
\begin{APACrefauthors}%
Bennewitz, M.%
, Burgard, W.%
\BCBL {} Thrun, S.%
\end{APACrefauthors}%
\unskip\
\newblock
\APACrefYearMonthDay{2002}{}{}.
\newblock
{\BBOQ}\APACrefatitle {Learning motion patterns of persons for mobile service
  robots} {Learning motion patterns of persons for mobile service
  robots}.{\BBCQ}
\newblock
 \APACrefbtitle {Proceedings 2002 IEEE International Conference on Robotics and
  Automation (Cat. No. 02CH37292)} {Proceedings 2002 ieee international
  conference on robotics and automation (cat. no. 02ch37292)}\ (\BVOL~4, \BPGS\
  3601--3606).
\PrintBackRefs{\CurrentBib}

\bibitem [\protect \citeauthoryear {%
Blundell%
, Cornebise%
, Kavukcuoglu%
\BCBL {}\ \BBA {} Wierstra%
}{%
Blundell%
\ \protect \BOthers {.}}{%
{\protect \APACyear {2015}}%
}]{%
blundell2015weight}
\APACinsertmetastar {%
blundell2015weight}%
\begin{APACrefauthors}%
Blundell, C.%
, Cornebise, J.%
, Kavukcuoglu, K.%
\BCBL {} Wierstra, D.%
\end{APACrefauthors}%
\unskip\
\newblock
\APACrefYearMonthDay{2015}{}{}.
\newblock
{\BBOQ}\APACrefatitle {Weight uncertainty in neural network} {Weight
  uncertainty in neural network}.{\BBCQ}
\newblock
 \APACrefbtitle {International conference on machine learning} {International
  conference on machine learning}\ (\BPGS\ 1613--1622).
\PrintBackRefs{\CurrentBib}

\bibitem [\protect \citeauthoryear {%
Deo%
\ \BBA {} Trivedi%
}{%
Deo%
\ \BBA {} Trivedi%
}{%
{\protect \APACyear {2020}}%
}]{%
deo2020trajectory}
\APACinsertmetastar {%
deo2020trajectory}%
\begin{APACrefauthors}%
Deo, N.%
\BCBT {}\ \BBA {} Trivedi, M.M.%
\end{APACrefauthors}%
\unskip\
\newblock
\APACrefYearMonthDay{2020}{}{}.
\newblock
{\BBOQ}\APACrefatitle {Trajectory forecasts in unknown environments conditioned
  on grid-based plans} {Trajectory forecasts in unknown environments
  conditioned on grid-based plans}.{\BBCQ}
\newblock
\APACjournalVolNumPages{arXiv preprint arXiv:2001.00735}{}{}{}.
\newblock

\newblock

\PrintBackRefs{\CurrentBib}

\bibitem [\protect \citeauthoryear {%
Dietrich%
\ \protect \BOthers {.}}{%
Dietrich%
\ \protect \BOthers {.}}{%
{\protect \APACyear {2021}}%
}]{%
dietrich2021learning}
\APACinsertmetastar {%
dietrich2021learning}%
\begin{APACrefauthors}%
Dietrich, F.%
, Makeev, A.%
, Kevrekidis, G.%
, Evangelou, N.%
, Bertalan, T.%
, Reich, S.%
\BCBL {} Kevrekidis, I.G.%
\end{APACrefauthors}%
\unskip\
\newblock
\APACrefYearMonthDay{2021}{}{}.
\newblock
{\BBOQ}\APACrefatitle {Learning effective stochastic differential equations
  from microscopic simulations: combining stochastic numerics and deep
  learning} {Learning effective stochastic differential equations from
  microscopic simulations: combining stochastic numerics and deep
  learning}.{\BBCQ}
\newblock
\APACjournalVolNumPages{arXiv preprint arXiv:2106.09004}{}{}{}.
\newblock

\newblock

\PrintBackRefs{\CurrentBib}

\bibitem [\protect \citeauthoryear {%
Ferrer%
\ \BBA {} Sanfeliu%
}{%
Ferrer%
\ \BBA {} Sanfeliu%
}{%
{\protect \APACyear {2014}}%
}]{%
ferrer2014behavior}
\APACinsertmetastar {%
ferrer2014behavior}%
\begin{APACrefauthors}%
Ferrer, G.%
\BCBT {}\ \BBA {} Sanfeliu, A.%
\end{APACrefauthors}%
\unskip\
\newblock
\APACrefYearMonthDay{2014}{}{}.
\newblock
{\BBOQ}\APACrefatitle {Behavior estimation for a complete framework for human
  motion prediction in crowded environments} {Behavior estimation for a
  complete framework for human motion prediction in crowded
  environments}.{\BBCQ}
\newblock
 \APACrefbtitle {2014 IEEE International Conference on Robotics and Automation
  (ICRA)} {2014 ieee international conference on robotics and automation
  (icra)}\ (\BPGS\ 5940--5945).
\PrintBackRefs{\CurrentBib}

\bibitem [\protect \citeauthoryear {%
Giuliari%
, Hasan%
, Cristani%
\BCBL {}\ \BBA {} Galasso%
}{%
Giuliari%
\ \protect \BOthers {.}}{%
{\protect \APACyear {2021}}%
}]{%
giuliari2021transformer}
\APACinsertmetastar {%
giuliari2021transformer}%
\begin{APACrefauthors}%
Giuliari, F.%
, Hasan, I.%
, Cristani, M.%
\BCBL {} Galasso, F.%
\end{APACrefauthors}%
\unskip\
\newblock
\APACrefYearMonthDay{2021}{}{}.
\newblock
{\BBOQ}\APACrefatitle {Transformer networks for trajectory forecasting}
  {Transformer networks for trajectory forecasting}.{\BBCQ}
\newblock
 \APACrefbtitle {2020 25th international conference on pattern recognition
  (ICPR)} {2020 25th international conference on pattern recognition (icpr)}\
  (\BPGS\ 10335--10342).
\PrintBackRefs{\CurrentBib}

\bibitem [\protect \citeauthoryear {%
Gong%
, Zhu%
, Andrew%
\BCBL {}\ \BBA {} Wang%
}{%
Gong%
\ \protect \BOthers {.}}{%
{\protect \APACyear {2022}}%
}]{%
Gong_finegrained_2022}
\APACinsertmetastar {%
Gong_finegrained_2022}%
\begin{APACrefauthors}%
Gong, D.%
, Zhu, Z.%
, Andrew, B.%
\BCBL {} Wang, H.%
\end{APACrefauthors}%
\unskip\
\newblock
\APACrefYearMonthDay{2022}{}{}.
\newblock
{\BBOQ}\APACrefatitle {Fine-grained differentiable physics: a yarn-level model
  for fabrics} {Fine-grained differentiable physics: a yarn-level model for
  fabrics}.{\BBCQ}
\newblock
 \APACrefbtitle {International Conference on Learning Representations.}
  {International conference on learning representations.}
\PrintBackRefs{\CurrentBib}

\bibitem [\protect \citeauthoryear {%
Goodfellow%
\ \protect \BOthers {.}}{%
Goodfellow%
\ \protect \BOthers {.}}{%
{\protect \APACyear {2020}}%
}]{%
goodfellow2020generative}
\APACinsertmetastar {%
goodfellow2020generative}%
\begin{APACrefauthors}%
Goodfellow, I.%
, Pouget-Abadie, J.%
, Mirza, M.%
, Xu, B.%
, Warde-Farley, D.%
, Ozair, S.%
\BDBL {}Bengio, Y.%
\end{APACrefauthors}%
\unskip\
\newblock
\APACrefYearMonthDay{2020}{}{}.
\newblock
{\BBOQ}\APACrefatitle {Generative adversarial networks} {Generative adversarial
  networks}.{\BBCQ}
\newblock
\APACjournalVolNumPages{Communications of the ACM}{63}{11}{139--144}.
\newblock

\newblock

\PrintBackRefs{\CurrentBib}

\bibitem [\protect \citeauthoryear {%
Gupta%
, Johnson%
, Fei-Fei%
, Savarese%
\BCBL {}\ \BBA {} Alahi%
}{%
Gupta%
\ \protect \BOthers {.}}{%
{\protect \APACyear {2018}}%
}]{%
gupta2018social}
\APACinsertmetastar {%
gupta2018social}%
\begin{APACrefauthors}%
Gupta, A.%
, Johnson, J.%
, Fei-Fei, L.%
, Savarese, S.%
\BCBL {} Alahi, A.%
\end{APACrefauthors}%
\unskip\
\newblock
\APACrefYearMonthDay{2018}{}{}.
\newblock
{\BBOQ}\APACrefatitle {Social gan: Socially acceptable trajectories with
  generative adversarial networks} {Social gan: Socially acceptable
  trajectories with generative adversarial networks}.{\BBCQ}
\newblock
 \APACrefbtitle {Proceedings of the IEEE conference on computer vision and
  pattern recognition} {Proceedings of the ieee conference on computer vision
  and pattern recognition}\ (\BPGS\ 2255--2264).
\PrintBackRefs{\CurrentBib}

\bibitem [\protect \citeauthoryear {%
He%
, Xia%
, Zhao%
\BCBL {}\ \BBA {} Wang%
}{%
He%
\ \protect \BOthers {.}}{%
{\protect \APACyear {2020}}%
}]{%
He_Informative_2020}
\APACinsertmetastar {%
He_Informative_2020}%
\begin{APACrefauthors}%
He, F.%
, Xia, Y.%
, Zhao, X.%
\BCBL {} Wang, H.%
\end{APACrefauthors}%
\unskip\
\newblock
\APACrefYearMonthDay{2020}{}{}.
\newblock
{\BBOQ}\APACrefatitle {Informative Scene Decomposition for Crowd Analysis,
  Comparison and Simulation Guidance} {Informative scene decomposition for
  crowd analysis, comparison and simulation guidance}.{\BBCQ}
\newblock
\APACjournalVolNumPages{ACM Transaction on Graphics (TOG)}{4}{39}{}.
\newblock

\newblock

\PrintBackRefs{\CurrentBib}

\bibitem [\protect \citeauthoryear {%
Helbing%
\ \BBA {} Moln\'ar%
}{%
Helbing%
\ \BBA {} Moln\'ar%
}{%
{\protect \APACyear {1995}}%
}]{%
Helbing_Social_1995}
\APACinsertmetastar {%
Helbing_Social_1995}%
\begin{APACrefauthors}%
Helbing, D.%
\BCBT {}\ \BBA {} Moln\'ar, P.%
\end{APACrefauthors}%
\unskip\
\newblock
\APACrefYearMonthDay{1995}{}{}.
\newblock
{\BBOQ}\APACrefatitle {Social force model for pedestrian dynamics} {Social
  force model for pedestrian dynamics}.{\BBCQ}
\newblock
\APACjournalVolNumPages{Phys. Rev. E}{51}{}{4282--4286}.
\newblock

\newblock

\PrintBackRefs{\CurrentBib}

\bibitem [\protect \citeauthoryear {%
Hochreiter%
\ \BBA {} Schmidhuber%
}{%
Hochreiter%
\ \BBA {} Schmidhuber%
}{%
{\protect \APACyear {1997}}%
}]{%
hochreiter1997long}
\APACinsertmetastar {%
hochreiter1997long}%
\begin{APACrefauthors}%
Hochreiter, S.%
\BCBT {}\ \BBA {} Schmidhuber, J.%
\end{APACrefauthors}%
\unskip\
\newblock
\APACrefYearMonthDay{1997}{}{}.
\newblock
{\BBOQ}\APACrefatitle {Long short-term memory} {Long short-term memory}.{\BBCQ}
\newblock
\APACjournalVolNumPages{Neural computation}{9}{8}{1735--1780}.
\newblock

\newblock

\PrintBackRefs{\CurrentBib}

\bibitem [\protect \citeauthoryear {%
Hossain%
, Johora%
, Müller%
, Hartmann%
\BCBL {}\ \BBA {} Reinhardt%
}{%
Hossain%
\ \protect \BOthers {.}}{%
{\protect \APACyear {2022}}%
}]{%
hossain22sfmgnet}
\APACinsertmetastar {%
hossain22sfmgnet}%
\begin{APACrefauthors}%
Hossain, S.%
, Johora, F.T.%
, Müller, J.P.%
, Hartmann, S.%
\BCBL {} Reinhardt, A.%
\end{APACrefauthors}%
\unskip\
\newblock
\APACrefYearMonthDay{2022}{}{}.
\newblock
{\BBOQ}\APACrefatitle {{SFMGNet: A Physics-based Neural Network To Predict
  Pedestrian Trajectories}} {{SFMGNet: A Physics-based Neural Network To
  Predict Pedestrian Trajectories}}.{\BBCQ}
\newblock
 \APACrefbtitle {Proceedings of the AAAI Spring Symposium on Machine Learning
  and Knowledge Engineering for Hybrid Intelligence (AAAI-MAKE)} {Proceedings
  of the aaai spring symposium on machine learning and knowledge engineering
  for hybrid intelligence (aaai-make)}\ (\BPGS\ 1--16).
\newblock
\begin{APACrefURL}
  {https://proceedings.aaai-make.info/AAAI-MAKE-PROCEEDINGS-2022/paper14.pdf}
  \end{APACrefURL}
\PrintBackRefs{\CurrentBib}

\bibitem [\protect \citeauthoryear {%
Huang%
, McGill%
, Williams%
, Fletcher%
\BCBL {}\ \BBA {} Rosman%
}{%
Huang%
\ \protect \BOthers {.}}{%
{\protect \APACyear {2019}}%
}]{%
huang2019uncertainty}
\APACinsertmetastar {%
huang2019uncertainty}%
\begin{APACrefauthors}%
Huang, X.%
, McGill, S.G.%
, Williams, B.C.%
, Fletcher, L.%
\BCBL {} Rosman, G.%
\end{APACrefauthors}%
\unskip\
\newblock
\APACrefYearMonthDay{2019}{}{}.
\newblock
{\BBOQ}\APACrefatitle {Uncertainty-aware driver trajectory prediction at urban
  intersections} {Uncertainty-aware driver trajectory prediction at urban
  intersections}.{\BBCQ}
\newblock
 \APACrefbtitle {2019 International conference on robotics and automation
  (ICRA)} {2019 international conference on robotics and automation (icra)}\
  (\BPGS\ 9718--9724).
\PrintBackRefs{\CurrentBib}

\bibitem [\protect \citeauthoryear {%
Ivanovic%
\ \BBA {} Pavone%
}{%
Ivanovic%
\ \BBA {} Pavone%
}{%
{\protect \APACyear {2019}}%
}]{%
ivanovic2019trajectron}
\APACinsertmetastar {%
ivanovic2019trajectron}%
\begin{APACrefauthors}%
Ivanovic, B.%
\BCBT {}\ \BBA {} Pavone, M.%
\end{APACrefauthors}%
\unskip\
\newblock
\APACrefYearMonthDay{2019}{}{}.
\newblock
{\BBOQ}\APACrefatitle {The trajectron: Probabilistic multi-agent trajectory
  modeling with dynamic spatiotemporal graphs} {The trajectron: Probabilistic
  multi-agent trajectory modeling with dynamic spatiotemporal graphs}.{\BBCQ}
\newblock
 \APACrefbtitle {Proceedings of the IEEE/CVF International Conference on
  Computer Vision} {Proceedings of the ieee/cvf international conference on
  computer vision}\ (\BPGS\ 2375--2384).
\PrintBackRefs{\CurrentBib}

\bibitem [\protect \citeauthoryear {%
Jospin%
, Laga%
, Boussaid%
, Buntine%
\BCBL {}\ \BBA {} Bennamoun%
}{%
Jospin%
\ \protect \BOthers {.}}{%
{\protect \APACyear {2022}}%
}]{%
jospin2022hands}
\APACinsertmetastar {%
jospin2022hands}%
\begin{APACrefauthors}%
Jospin, L.V.%
, Laga, H.%
, Boussaid, F.%
, Buntine, W.%
\BCBL {} Bennamoun, M.%
\end{APACrefauthors}%
\unskip\
\newblock
\APACrefYearMonthDay{2022}{}{}.
\newblock
{\BBOQ}\APACrefatitle {Hands-on Bayesian neural networks—A tutorial for deep
  learning users} {Hands-on bayesian neural networks—a tutorial for deep
  learning users}.{\BBCQ}
\newblock
\APACjournalVolNumPages{IEEE Computational Intelligence
  Magazine}{17}{2}{29--48}.
\newblock

\newblock

\PrintBackRefs{\CurrentBib}

\bibitem [\protect \citeauthoryear {%
Karniadakis%
\ \protect \BOthers {.}}{%
Karniadakis%
\ \protect \BOthers {.}}{%
{\protect \APACyear {2021}}%
}]{%
karniadakis2021physics}
\APACinsertmetastar {%
karniadakis2021physics}%
\begin{APACrefauthors}%
Karniadakis, G.E.%
, Kevrekidis, I.G.%
, Lu, L.%
, Perdikaris, P.%
, Wang, S.%
\BCBL {} Yang, L.%
\end{APACrefauthors}%
\unskip\
\newblock
\APACrefYearMonthDay{2021}{}{}.
\newblock
{\BBOQ}\APACrefatitle {Physics-informed machine learning} {Physics-informed
  machine learning}.{\BBCQ}
\newblock
\APACjournalVolNumPages{Nature Reviews Physics}{3}{6}{422--440}.
\newblock

\newblock

\PrintBackRefs{\CurrentBib}

\bibitem [\protect \citeauthoryear {%
Kim%
\ \protect \BOthers {.}}{%
Kim%
\ \protect \BOthers {.}}{%
{\protect \APACyear {2013}}%
}]{%
kim2013predicting}
\APACinsertmetastar {%
kim2013predicting}%
\begin{APACrefauthors}%
Kim, S.%
, Guy, S.J.%
, Liu, W.%
, Lau, R.W.%
, Lin, M.C.%
\BCBL {} Manocha, D.%
\end{APACrefauthors}%
\unskip\
\newblock
\APACrefYearMonthDay{2013}{}{}.
\newblock
{\BBOQ}\APACrefatitle {Predicting pedestrian trajectories using velocity-space
  reasoning} {Predicting pedestrian trajectories using velocity-space
  reasoning}.{\BBCQ}
\newblock
 \APACrefbtitle {Algorithmic foundations of robotics X} {Algorithmic
  foundations of robotics x}\ (\BPGS\ 609--623).
\newblock
\APACaddressPublisher{}{Springer}.
\PrintBackRefs{\CurrentBib}

\bibitem [\protect \citeauthoryear {%
Kingma%
, Welling%
\BCBL {}\ \protect \BOthers {.}}{%
Kingma%
\ \protect \BOthers {.}}{%
{\protect \APACyear {2019}}%
}]{%
kingma2019introduction}
\APACinsertmetastar {%
kingma2019introduction}%
\begin{APACrefauthors}%
Kingma, D.P.%
, Welling, M.%
\BCBL {}\ \BOthersPeriod {.}\end{APACrefauthors}%
\unskip\
\newblock
\APACrefYearMonthDay{2019}{}{}.
\newblock
{\BBOQ}\APACrefatitle {An introduction to variational autoencoders} {An
  introduction to variational autoencoders}.{\BBCQ}
\newblock
\APACjournalVolNumPages{Foundations and Trends{\textregistered} in Machine
  Learning}{12}{4}{307--392}.
\newblock

\newblock

\PrintBackRefs{\CurrentBib}

\bibitem [\protect \citeauthoryear {%
Kosaraju%
\ \protect \BOthers {.}}{%
Kosaraju%
\ \protect \BOthers {.}}{%
{\protect \APACyear {2019}}%
}]{%
kosaraju2019social}
\APACinsertmetastar {%
kosaraju2019social}%
\begin{APACrefauthors}%
Kosaraju, V.%
, Sadeghian, A.%
, Mart{\'\i}n-Mart{\'\i}n, R.%
, Reid, I.%
, Rezatofighi, H.%
\BCBL {} Savarese, S.%
\end{APACrefauthors}%
\unskip\
\newblock
\APACrefYearMonthDay{2019}{}{}.
\newblock
{\BBOQ}\APACrefatitle {Social-bigat: Multimodal trajectory forecasting using
  bicycle-gan and graph attention networks} {Social-bigat: Multimodal
  trajectory forecasting using bicycle-gan and graph attention
  networks}.{\BBCQ}
\newblock
\APACjournalVolNumPages{Advances in Neural Information Processing
  Systems}{32}{}{}.
\newblock

\newblock

\PrintBackRefs{\CurrentBib}

\bibitem [\protect \citeauthoryear {%
Kreiss%
}{%
Kreiss%
}{%
{\protect \APACyear {2021}}%
}]{%
kreiss2021deep}
\APACinsertmetastar {%
kreiss2021deep}%
\begin{APACrefauthors}%
Kreiss, S.%
\end{APACrefauthors}%
\unskip\
\newblock
\APACrefYearMonthDay{2021}{}{}.
\newblock
{\BBOQ}\APACrefatitle {Deep Social Force} {Deep social force}.{\BBCQ}
\newblock
\APACjournalVolNumPages{arXiv preprint arXiv:2109.12081}{}{}{}.
\newblock

\newblock

\PrintBackRefs{\CurrentBib}

\bibitem [\protect \citeauthoryear {%
Lerner%
, Chrysanthou%
\BCBL {}\ \BBA {} Lischinski%
}{%
Lerner%
\ \protect \BOthers {.}}{%
{\protect \APACyear {2007}}%
}]{%
lerner2007crowds}
\APACinsertmetastar {%
lerner2007crowds}%
\begin{APACrefauthors}%
Lerner, A.%
, Chrysanthou, Y.%
\BCBL {} Lischinski, D.%
\end{APACrefauthors}%
\unskip\
\newblock
\APACrefYearMonthDay{2007}{}{}.
\newblock
{\BBOQ}\APACrefatitle {Crowds by example} {Crowds by example}.{\BBCQ}
\newblock
 \APACrefbtitle {Computer graphics forum} {Computer graphics forum}\ (\BVOL~26,
  \BPGS\ 655--664).
\PrintBackRefs{\CurrentBib}

\bibitem [\protect \citeauthoryear {%
Li%
, Wong%
, Chen%
\BCBL {}\ \BBA {} Duvenaud%
}{%
Li%
\ \protect \BOthers {.}}{%
{\protect \APACyear {2020}}%
}]{%
li2020scalable}
\APACinsertmetastar {%
li2020scalable}%
\begin{APACrefauthors}%
Li, X.%
, Wong, T\BHBI K.L.%
, Chen, R.T.%
\BCBL {} Duvenaud, D.%
\end{APACrefauthors}%
\unskip\
\newblock
\APACrefYearMonthDay{2020}{}{}.
\newblock
{\BBOQ}\APACrefatitle {Scalable Gradients for Stochastic Differential
  Equations} {Scalable gradients for stochastic differential equations}.{\BBCQ}
\newblock
 \APACrefbtitle {International Conference on Artificial Intelligence and
  Statistics} {International conference on artificial intelligence and
  statistics}\ (\BPGS\ 3870--3882).
\PrintBackRefs{\CurrentBib}

\bibitem [\protect \citeauthoryear {%
Liang%
, Jiang%
\BCBL {}\ \BBA {} Hauptmann%
}{%
Liang%
\ \protect \BOthers {.}}{%
{\protect \APACyear {2020}}%
}]{%
liang2020simaug}
\APACinsertmetastar {%
liang2020simaug}%
\begin{APACrefauthors}%
Liang, J.%
, Jiang, L.%
\BCBL {} Hauptmann, A.%
\end{APACrefauthors}%
\unskip\
\newblock
\APACrefYearMonthDay{2020}{}{}.
\newblock
{\BBOQ}\APACrefatitle {Simaug: Learning robust representations from 3d
  simulation for pedestrian trajectory prediction in unseen cameras} {Simaug:
  Learning robust representations from 3d simulation for pedestrian trajectory
  prediction in unseen cameras}.{\BBCQ}
\newblock
\APACjournalVolNumPages{arXiv preprint arXiv:2004.02022}{2}{}{}.
\newblock

\newblock

\PrintBackRefs{\CurrentBib}

\bibitem [\protect \citeauthoryear {%
Liang%
, Jiang%
, Niebles%
, Hauptmann%
\BCBL {}\ \BBA {} Fei-Fei%
}{%
Liang%
\ \protect \BOthers {.}}{%
{\protect \APACyear {2019}}%
}]{%
liang2019peeking}
\APACinsertmetastar {%
liang2019peeking}%
\begin{APACrefauthors}%
Liang, J.%
, Jiang, L.%
, Niebles, J.C.%
, Hauptmann, A.G.%
\BCBL {} Fei-Fei, L.%
\end{APACrefauthors}%
\unskip\
\newblock
\APACrefYearMonthDay{2019}{}{}.
\newblock
{\BBOQ}\APACrefatitle {Peeking into the future: Predicting future person
  activities and locations in videos} {Peeking into the future: Predicting
  future person activities and locations in videos}.{\BBCQ}
\newblock
 \APACrefbtitle {Proceedings of the IEEE/CVF Conference on Computer Vision and
  Pattern Recognition} {Proceedings of the ieee/cvf conference on computer
  vision and pattern recognition}\ (\BPGS\ 5725--5734).
\PrintBackRefs{\CurrentBib}

\bibitem [\protect \citeauthoryear {%
Luo%
\ \protect \BOthers {.}}{%
Luo%
\ \protect \BOthers {.}}{%
{\protect \APACyear {2008}}%
}]{%
luo2008agent}
\APACinsertmetastar {%
luo2008agent}%
\begin{APACrefauthors}%
Luo, L.%
, Zhou, S.%
, Cai, W.%
, Low, M.Y.H.%
, Tian, F.%
, Wang, Y.%
\BDBL {}Chen, D.%
\end{APACrefauthors}%
\unskip\
\newblock
\APACrefYearMonthDay{2008}{}{}.
\newblock
{\BBOQ}\APACrefatitle {Agent-based human behavior modeling for crowd
  simulation} {Agent-based human behavior modeling for crowd
  simulation}.{\BBCQ}
\newblock
\APACjournalVolNumPages{Computer Animation and Virtual
  Worlds}{19}{3-4}{271--281}.
\newblock

\newblock

\PrintBackRefs{\CurrentBib}

\bibitem [\protect \citeauthoryear {%
Mangalam%
, An%
, Girase%
\BCBL {}\ \BBA {} Malik%
}{%
Mangalam%
\ \protect \BOthers {.}}{%
{\protect \APACyear {2021}}%
}]{%
mangalam2021goals}
\APACinsertmetastar {%
mangalam2021goals}%
\begin{APACrefauthors}%
Mangalam, K.%
, An, Y.%
, Girase, H.%
\BCBL {} Malik, J.%
\end{APACrefauthors}%
\unskip\
\newblock
\APACrefYearMonthDay{2021}{}{}.
\newblock
{\BBOQ}\APACrefatitle {From goals, waypoints \& paths to long term human
  trajectory forecasting} {From goals, waypoints \& paths to long term human
  trajectory forecasting}.{\BBCQ}
\newblock
 \APACrefbtitle {Proceedings of the IEEE/CVF International Conference on
  Computer Vision} {Proceedings of the ieee/cvf international conference on
  computer vision}\ (\BPGS\ 15233--15242).
\PrintBackRefs{\CurrentBib}

\bibitem [\protect \citeauthoryear {%
Mangalam%
\ \protect \BOthers {.}}{%
Mangalam%
\ \protect \BOthers {.}}{%
{\protect \APACyear {2020}}%
}]{%
mangalam2020not}
\APACinsertmetastar {%
mangalam2020not}%
\begin{APACrefauthors}%
Mangalam, K.%
, Girase, H.%
, Agarwal, S.%
, Lee, K\BHBI H.%
, Adeli, E.%
, Malik, J.%
\BCBL {} Gaidon, A.%
\end{APACrefauthors}%
\unskip\
\newblock
\APACrefYearMonthDay{2020}{}{}.
\newblock
{\BBOQ}\APACrefatitle {It is not the journey but the destination: Endpoint
  conditioned trajectory prediction} {It is not the journey but the
  destination: Endpoint conditioned trajectory prediction}.{\BBCQ}
\newblock
 \APACrefbtitle {European Conference on Computer Vision} {European conference
  on computer vision}\ (\BPGS\ 759--776).
\PrintBackRefs{\CurrentBib}

\bibitem [\protect \citeauthoryear {%
Mohamed%
, Qian%
, Elhoseiny%
\BCBL {}\ \BBA {} Claudel%
}{%
Mohamed%
\ \protect \BOthers {.}}{%
{\protect \APACyear {2020}}%
}]{%
mohamed2020social}
\APACinsertmetastar {%
mohamed2020social}%
\begin{APACrefauthors}%
Mohamed, A.%
, Qian, K.%
, Elhoseiny, M.%
\BCBL {} Claudel, C.%
\end{APACrefauthors}%
\unskip\
\newblock
\APACrefYearMonthDay{2020}{}{}.
\newblock
{\BBOQ}\APACrefatitle {Social-stgcnn: A social spatio-temporal graph
  convolutional neural network for human trajectory prediction} {Social-stgcnn:
  A social spatio-temporal graph convolutional neural network for human
  trajectory prediction}.{\BBCQ}
\newblock
 \APACrefbtitle {Proceedings of the IEEE/CVF Conference on Computer Vision and
  Pattern Recognition} {Proceedings of the ieee/cvf conference on computer
  vision and pattern recognition}\ (\BPGS\ 14424--14432).
\PrintBackRefs{\CurrentBib}

\bibitem [\protect \citeauthoryear {%
Narang%
, Best%
, Curtis%
\BCBL {}\ \BBA {} Manocha%
}{%
Narang%
\ \protect \BOthers {.}}{%
{\protect \APACyear {2015}}%
}]{%
narang2015generating}
\APACinsertmetastar {%
narang2015generating}%
\begin{APACrefauthors}%
Narang, S.%
, Best, A.%
, Curtis, S.%
\BCBL {} Manocha, D.%
\end{APACrefauthors}%
\unskip\
\newblock
\APACrefYearMonthDay{2015}{}{}.
\newblock
{\BBOQ}\APACrefatitle {Generating pedestrian trajectories consistent with the
  fundamental diagram based on physiological and psychological factors}
  {Generating pedestrian trajectories consistent with the fundamental diagram
  based on physiological and psychological factors}.{\BBCQ}
\newblock
\APACjournalVolNumPages{PLoS one}{10}{4}{e0117856}.
\newblock

\newblock

\PrintBackRefs{\CurrentBib}

\bibitem [\protect \citeauthoryear {%
Opper%
}{%
Opper%
}{%
{\protect \APACyear {2019}}%
}]{%
opper2019variational}
\APACinsertmetastar {%
opper2019variational}%
\begin{APACrefauthors}%
Opper, M.%
\end{APACrefauthors}%
\unskip\
\newblock
\APACrefYearMonthDay{2019}{}{}.
\newblock
{\BBOQ}\APACrefatitle {Variational inference for stochastic differential
  equations} {Variational inference for stochastic differential
  equations}.{\BBCQ}
\newblock
\APACjournalVolNumPages{Annalen der Physik}{531}{3}{1800233}.
\newblock

\newblock

\PrintBackRefs{\CurrentBib}

\bibitem [\protect \citeauthoryear {%
Pellegrini%
, Ess%
\BCBL {}\ \BBA {} Gool%
}{%
Pellegrini%
\ \protect \BOthers {.}}{%
{\protect \APACyear {2010}}%
}]{%
pellegrini2010improving}
\APACinsertmetastar {%
pellegrini2010improving}%
\begin{APACrefauthors}%
Pellegrini, S.%
, Ess, A.%
\BCBL {} Gool, L.V.%
\end{APACrefauthors}%
\unskip\
\newblock
\APACrefYearMonthDay{2010}{}{}.
\newblock
{\BBOQ}\APACrefatitle {Improving data association by joint modeling of
  pedestrian trajectories and groupings} {Improving data association by joint
  modeling of pedestrian trajectories and groupings}.{\BBCQ}
\newblock
 \APACrefbtitle {European conference on computer vision} {European conference
  on computer vision}\ (\BPGS\ 452--465).
\PrintBackRefs{\CurrentBib}

\bibitem [\protect \citeauthoryear {%
Robicquet%
, Sadeghian%
, Alahi%
\BCBL {}\ \BBA {} Savarese%
}{%
Robicquet%
\ \protect \BOthers {.}}{%
{\protect \APACyear {2016}}%
}]{%
robicquet2016learning}
\APACinsertmetastar {%
robicquet2016learning}%
\begin{APACrefauthors}%
Robicquet, A.%
, Sadeghian, A.%
, Alahi, A.%
\BCBL {} Savarese, S.%
\end{APACrefauthors}%
\unskip\
\newblock
\APACrefYearMonthDay{2016}{}{}.
\newblock
{\BBOQ}\APACrefatitle {Learning social etiquette: Human trajectory
  understanding in crowded scenes} {Learning social etiquette: Human trajectory
  understanding in crowded scenes}.{\BBCQ}
\newblock
 \APACrefbtitle {European conference on computer vision} {European conference
  on computer vision}\ (\BPGS\ 549--565).
\PrintBackRefs{\CurrentBib}

\bibitem [\protect \citeauthoryear {%
Sadeghian%
\ \protect \BOthers {.}}{%
Sadeghian%
\ \protect \BOthers {.}}{%
{\protect \APACyear {2019}}%
}]{%
sadeghian2019sophie}
\APACinsertmetastar {%
sadeghian2019sophie}%
\begin{APACrefauthors}%
Sadeghian, A.%
, Kosaraju, V.%
, Sadeghian, A.%
, Hirose, N.%
, Rezatofighi, H.%
\BCBL {} Savarese, S.%
\end{APACrefauthors}%
\unskip\
\newblock
\APACrefYearMonthDay{2019}{}{}.
\newblock
{\BBOQ}\APACrefatitle {Sophie: An attentive gan for predicting paths compliant
  to social and physical constraints} {Sophie: An attentive gan for predicting
  paths compliant to social and physical constraints}.{\BBCQ}
\newblock
 \APACrefbtitle {Proceedings of the IEEE/CVF Conference on Computer Vision and
  Pattern Recognition} {Proceedings of the ieee/cvf conference on computer
  vision and pattern recognition}\ (\BPGS\ 1349--1358).
\PrintBackRefs{\CurrentBib}

\bibitem [\protect \citeauthoryear {%
Schneider%
\ \BBA {} Gavrila%
}{%
Schneider%
\ \BBA {} Gavrila%
}{%
{\protect \APACyear {2013}}%
}]{%
schneider2013pedestrian}
\APACinsertmetastar {%
schneider2013pedestrian}%
\begin{APACrefauthors}%
Schneider, N.%
\BCBT {}\ \BBA {} Gavrila, D.M.%
\end{APACrefauthors}%
\unskip\
\newblock
\APACrefYearMonthDay{2013}{}{}.
\newblock
{\BBOQ}\APACrefatitle {Pedestrian path prediction with recursive bayesian
  filters: A comparative study} {Pedestrian path prediction with recursive
  bayesian filters: A comparative study}.{\BBCQ}
\newblock
 \APACrefbtitle {german conference on pattern recognition} {german conference
  on pattern recognition}\ (\BPGS\ 174--183).
\PrintBackRefs{\CurrentBib}

\bibitem [\protect \citeauthoryear {%
Shen%
\ \protect \BOthers {.}}{%
Shen%
\ \protect \BOthers {.}}{%
{\protect \APACyear {2021}}%
}]{%
Shen_high_2021}
\APACinsertmetastar {%
Shen_high_2021}%
\begin{APACrefauthors}%
Shen, S.%
, Yang, Y.%
, Shao, T.%
, Wang, H.%
, Jiang, C.%
, Lan, L.%
\BCBL {} Zhou, K.%
\end{APACrefauthors}%
\unskip\
\newblock
\APACrefYearMonthDay{2021}{jul}{}.
\newblock
{\BBOQ}\APACrefatitle {High-Order Differentiable Autoencoder for Nonlinear
  Model Reduction} {High-order differentiable autoencoder for nonlinear model
  reduction}.{\BBCQ}
\newblock
\APACjournalVolNumPages{ACM Trans. Graph.}{40}{4}{}.
\newblock

\newblock

\PrintBackRefs{\CurrentBib}

\bibitem [\protect \citeauthoryear {%
Shi%
\ \protect \BOthers {.}}{%
Shi%
\ \protect \BOthers {.}}{%
{\protect \APACyear {2021}}%
}]{%
shi2021sgcn}
\APACinsertmetastar {%
shi2021sgcn}%
\begin{APACrefauthors}%
Shi, L.%
, Wang, L.%
, Long, C.%
, Zhou, S.%
, Zhou, M.%
, Niu, Z.%
\BCBL {} Hua, G.%
\end{APACrefauthors}%
\unskip\
\newblock
\APACrefYearMonthDay{2021}{}{}.
\newblock
{\BBOQ}\APACrefatitle {SGCN: Sparse graph convolution network for pedestrian
  trajectory prediction} {Sgcn: Sparse graph convolution network for pedestrian
  trajectory prediction}.{\BBCQ}
\newblock
 \APACrefbtitle {Proceedings of the IEEE/CVF Conference on Computer Vision and
  Pattern Recognition} {Proceedings of the ieee/cvf conference on computer
  vision and pattern recognition}\ (\BPGS\ 8994--9003).
\PrintBackRefs{\CurrentBib}

\bibitem [\protect \citeauthoryear {%
Thrun%
, Burgard%
\BCBL {}\ \BBA {} Fox%
}{%
Thrun%
\ \protect \BOthers {.}}{%
{\protect \APACyear {2005}}%
}]{%
thrun2005probabilistic}
\APACinsertmetastar {%
thrun2005probabilistic}%
\begin{APACrefauthors}%
Thrun, S.%
, Burgard, W.%
\BCBL {} Fox, D.%
\end{APACrefauthors}%
\unskip\
\newblock
\APACrefYearMonthDay{2005}{}{}.
\newblock
\APACrefbtitle {Probabilistic Robotics (Intelligent Robotics and Autonomous
  Agents series), ser. Intelligent robotics and autonomous agents.}
  {Probabilistic robotics (intelligent robotics and autonomous agents series),
  ser. intelligent robotics and autonomous agents.}
\newblock
\APACaddressPublisher{}{The MIT Press}.
\PrintBackRefs{\CurrentBib}

\bibitem [\protect \citeauthoryear {%
Tzen%
\ \BBA {} Raginsky%
}{%
Tzen%
\ \BBA {} Raginsky%
}{%
{\protect \APACyear {2019}}%
}]{%
tzen2019neural}
\APACinsertmetastar {%
tzen2019neural}%
\begin{APACrefauthors}%
Tzen, B.%
\BCBT {}\ \BBA {} Raginsky, M.%
\end{APACrefauthors}%
\unskip\
\newblock
\APACrefYearMonthDay{2019}{}{}.
\newblock
{\BBOQ}\APACrefatitle {Neural stochastic differential equations: Deep latent
  gaussian models in the diffusion limit} {Neural stochastic differential
  equations: Deep latent gaussian models in the diffusion limit}.{\BBCQ}
\newblock
\APACjournalVolNumPages{arXiv preprint arXiv:1905.09883}{}{}{}.
\newblock

\newblock

\PrintBackRefs{\CurrentBib}

\bibitem [\protect \citeauthoryear {%
Vemula%
, Muelling%
\BCBL {}\ \BBA {} Oh%
}{%
Vemula%
\ \protect \BOthers {.}}{%
{\protect \APACyear {2018}}%
}]{%
vemula2018social}
\APACinsertmetastar {%
vemula2018social}%
\begin{APACrefauthors}%
Vemula, A.%
, Muelling, K.%
\BCBL {} Oh, J.%
\end{APACrefauthors}%
\unskip\
\newblock
\APACrefYearMonthDay{2018}{}{}.
\newblock
{\BBOQ}\APACrefatitle {Social attention: Modeling attention in human crowds}
  {Social attention: Modeling attention in human crowds}.{\BBCQ}
\newblock
 \APACrefbtitle {2018 IEEE international Conference on Robotics and Automation
  (ICRA)} {2018 ieee international conference on robotics and automation
  (icra)}\ (\BPGS\ 4601--4607).
\PrintBackRefs{\CurrentBib}

\bibitem [\protect \citeauthoryear {%
Wang%
, Ond{\v{r}}ej%
\BCBL {}\ \BBA {} O'Sullivan%
}{%
Wang%
\ \protect \BOthers {.}}{%
{\protect \APACyear {2016}}%
{\protect \APACexlab {{\protect \BCnt {1}}}}}]{%
wang_path_2016}
\APACinsertmetastar {%
wang_path_2016}%
\begin{APACrefauthors}%
Wang, H.%
, Ond{\v{r}}ej, J.%
\BCBL {} O'Sullivan, C.%
\end{APACrefauthors}%
\unskip\
\newblock
\APACrefYearMonthDay{2016{\protect \BCnt {1}}}{}{}.
\newblock
{\BBOQ}\APACrefatitle {Path Patterns: Analyzing and Comparing Real and
  Simulated Crowds} {Path patterns: Analyzing and comparing real and simulated
  crowds}.{\BBCQ}
\newblock
 \APACrefbtitle {ACM SIGGRAPH Symposium on Interactive 3D Graphics and Games
  2016} {Acm siggraph symposium on interactive 3d graphics and games 2016}\
  (\BPGS\ 49--57).
\PrintBackRefs{\CurrentBib}

\bibitem [\protect \citeauthoryear {%
Wang%
, Ond{\v{r}}ej%
\BCBL {}\ \BBA {} O'Sullivan%
}{%
Wang%
\ \protect \BOthers {.}}{%
{\protect \APACyear {2016}}%
{\protect \APACexlab {{\protect \BCnt {2}}}}}]{%
wang_trending_2016}
\APACinsertmetastar {%
wang_trending_2016}%
\begin{APACrefauthors}%
Wang, H.%
, Ond{\v{r}}ej, J.%
\BCBL {} O'Sullivan, C.%
\end{APACrefauthors}%
\unskip\
\newblock
\APACrefYearMonthDay{2016{\protect \BCnt {2}}}{}{}.
\newblock
{\BBOQ}\APACrefatitle {Trending Paths: A New Semantic-level Metric for
  Comparing Simulated and Real Crowd Data} {Trending paths: A new
  semantic-level metric for comparing simulated and real crowd data}.{\BBCQ}
\newblock
\APACjournalVolNumPages{IEEE Transactions on Visualization and Computer
  Graphics}{PP}{99}{1-1}.
\newblock

\newblock

\PrintBackRefs{\CurrentBib}

\bibitem [\protect \citeauthoryear {%
Wang%
\ \BBA {} O’Sullivan%
}{%
Wang%
\ \BBA {} O’Sullivan%
}{%
{\protect \APACyear {2016}}%
}]{%
wang2016globally}
\APACinsertmetastar {%
wang2016globally}%
\begin{APACrefauthors}%
Wang, H.%
\BCBT {}\ \BBA {} O’Sullivan, C.%
\end{APACrefauthors}%
\unskip\
\newblock
\APACrefYearMonthDay{2016}{}{}.
\newblock
{\BBOQ}\APACrefatitle {Globally continuous and non-Markovian crowd activity
  analysis from videos} {Globally continuous and non-markovian crowd activity
  analysis from videos}.{\BBCQ}
\newblock
 \APACrefbtitle {European conference on computer vision} {European conference
  on computer vision}\ (\BPGS\ 527--544).
\PrintBackRefs{\CurrentBib}

\bibitem [\protect \citeauthoryear {%
Wong%
\ \protect \BOthers {.}}{%
Wong%
\ \protect \BOthers {.}}{%
{\protect \APACyear {2022}}%
}]{%
wong2022view}
\APACinsertmetastar {%
wong2022view}%
\begin{APACrefauthors}%
Wong, C.%
, Xia, B.%
, Hong, Z.%
, Peng, Q.%
, Yuan, W.%
, Cao, Q.%
\BDBL {}You, X.%
\end{APACrefauthors}%
\unskip\
\newblock
\APACrefYearMonthDay{2022}{}{}.
\newblock
{\BBOQ}\APACrefatitle {View Vertically: A hierarchical network for trajectory
  prediction via fourier spectrums} {View vertically: A hierarchical network
  for trajectory prediction via fourier spectrums}.{\BBCQ}
\newblock
 \APACrefbtitle {European Conference on Computer Vision} {European conference
  on computer vision}\ (\BPGS\ 682--700).
\PrintBackRefs{\CurrentBib}

\bibitem [\protect \citeauthoryear {%
Xu%
, Hayet%
\BCBL {}\ \BBA {} Karamouzas%
}{%
Xu%
\ \protect \BOthers {.}}{%
{\protect \APACyear {2022}}%
}]{%
xu2022socialvae}
\APACinsertmetastar {%
xu2022socialvae}%
\begin{APACrefauthors}%
Xu, P.%
, Hayet, J\BHBI B.%
\BCBL {} Karamouzas, I.%
\end{APACrefauthors}%
\unskip\
\newblock
\APACrefYearMonthDay{2022}{}{}.
\newblock
{\BBOQ}\APACrefatitle {SocialVAE: Human Trajectory Prediction using Timewise
  Latents} {Socialvae: Human trajectory prediction using timewise
  latents}.{\BBCQ}
\newblock
\APACjournalVolNumPages{arXiv preprint arXiv:2203.08207}{}{}{}.
\newblock

\newblock

\PrintBackRefs{\CurrentBib}

\bibitem [\protect \citeauthoryear {%
Yan%
, Kakadiaris%
\BCBL {}\ \BBA {} Shah%
}{%
Yan%
\ \protect \BOthers {.}}{%
{\protect \APACyear {2014}}%
}]{%
yan2014modeling}
\APACinsertmetastar {%
yan2014modeling}%
\begin{APACrefauthors}%
Yan, X.%
, Kakadiaris, I.A.%
\BCBL {} Shah, S.K.%
\end{APACrefauthors}%
\unskip\
\newblock
\APACrefYearMonthDay{2014}{}{}.
\newblock
{\BBOQ}\APACrefatitle {Modeling local behavior for predicting social
  interactions towards human tracking} {Modeling local behavior for predicting
  social interactions towards human tracking}.{\BBCQ}
\newblock
\APACjournalVolNumPages{Pattern Recognition}{47}{4}{1626--1641}.
\newblock

\newblock

\PrintBackRefs{\CurrentBib}

\bibitem [\protect \citeauthoryear {%
Yu%
, Ma%
, Ren%
, Zhao%
\BCBL {}\ \BBA {} Yi%
}{%
Yu%
\ \protect \BOthers {.}}{%
{\protect \APACyear {2020}}%
}]{%
yu2020spatio}
\APACinsertmetastar {%
yu2020spatio}%
\begin{APACrefauthors}%
Yu, C.%
, Ma, X.%
, Ren, J.%
, Zhao, H.%
\BCBL {} Yi, S.%
\end{APACrefauthors}%
\unskip\
\newblock
\APACrefYearMonthDay{2020}{}{}.
\newblock
{\BBOQ}\APACrefatitle {Spatio-temporal graph transformer networks for
  pedestrian trajectory prediction} {Spatio-temporal graph transformer networks
  for pedestrian trajectory prediction}.{\BBCQ}
\newblock
 \APACrefbtitle {European Conference on Computer Vision} {European conference
  on computer vision}\ (\BPGS\ 507--523).
\PrintBackRefs{\CurrentBib}

\bibitem [\protect \citeauthoryear {%
{Yue}%
, {Manocha}%
\BCBL {}\ \BBA {} {Wang}%
}{%
{Yue}%
\ \protect \BOthers {.}}{%
{\protect \APACyear {2022}}%
}]{%
Yue_trajectory_2022}
\APACinsertmetastar {%
Yue_trajectory_2022}%
\begin{APACrefauthors}%
{Yue}, J.%
, {Manocha}, D.%
\BCBL {} {Wang}, H.%
\end{APACrefauthors}%
\unskip\
\newblock
\APACrefYearMonthDay{2022}{}{}.
\newblock
{\BBOQ}\APACrefatitle {Human Trajectory Prediction via Neural Social Physics}
  {Human trajectory prediction via neural social physics}.{\BBCQ}
\newblock
 \APACrefbtitle {Proceedings of the European Conference on Computer Vision
  (ECCV).} {Proceedings of the european conference on computer vision (eccv).}
\PrintBackRefs{\CurrentBib}

\bibitem [\protect \citeauthoryear {%
P.~Zhang%
, Ouyang%
, Zhang%
, Xue%
\BCBL {}\ \BBA {} Zheng%
}{%
P.~Zhang%
\ \protect \BOthers {.}}{%
{\protect \APACyear {2019}}%
}]{%
zhang2019sr}
\APACinsertmetastar {%
zhang2019sr}%
\begin{APACrefauthors}%
Zhang, P.%
, Ouyang, W.%
, Zhang, P.%
, Xue, J.%
\BCBL {} Zheng, N.%
\end{APACrefauthors}%
\unskip\
\newblock
\APACrefYearMonthDay{2019}{}{}.
\newblock
{\BBOQ}\APACrefatitle {Sr-lstm: State refinement for lstm towards pedestrian
  trajectory prediction} {Sr-lstm: State refinement for lstm towards pedestrian
  trajectory prediction}.{\BBCQ}
\newblock
 \APACrefbtitle {Proceedings of the IEEE/CVF Conference on Computer Vision and
  Pattern Recognition} {Proceedings of the ieee/cvf conference on computer
  vision and pattern recognition}\ (\BPGS\ 12085--12094).
\PrintBackRefs{\CurrentBib}

\bibitem [\protect \citeauthoryear {%
Z.~Zhang%
, Jimack%
\BCBL {}\ \BBA {} Wang%
}{%
Z.~Zhang%
\ \protect \BOthers {.}}{%
{\protect \APACyear {2021}}%
}]{%
zhang_meshingnet3d_2021}
\APACinsertmetastar {%
zhang_meshingnet3d_2021}%
\begin{APACrefauthors}%
Zhang, Z.%
, Jimack, P.K.%
\BCBL {} Wang, H.%
\end{APACrefauthors}%
\unskip\
\newblock
\APACrefYearMonthDay{2021}{{\APACmonth{07}}}{}.
\newblock
{\BBOQ}\APACrefatitle {{MeshingNet3D}: {Efficient} generation of adapted
  tetrahedral meshes for computational mechanics} {{MeshingNet3D}: {Efficient}
  generation of adapted tetrahedral meshes for computational mechanics}.{\BBCQ}
\newblock
\APACjournalVolNumPages{Advances in Engineering Software}{157-158}{}{}.
\newblock

\newblock

\PrintBackRefs{\CurrentBib}

\bibitem [\protect \citeauthoryear {%
Z.~Zhang%
, Wang%
, Jimack%
\BCBL {}\ \BBA {} Wang%
}{%
Z.~Zhang%
\ \protect \BOthers {.}}{%
{\protect \APACyear {2020}}%
}]{%
Zhang_MeshingNet_2020}
\APACinsertmetastar {%
Zhang_MeshingNet_2020}%
\begin{APACrefauthors}%
Zhang, Z.%
, Wang, Y.%
, Jimack, P.K.%
\BCBL {} Wang, H.%
\end{APACrefauthors}%
\unskip\
\newblock
\APACrefYearMonthDay{2020}{}{}.
\newblock
{\BBOQ}\APACrefatitle {MeshingNet: A New Mesh Generation Method Based on Deep
  Learning} {Meshingnet: A new mesh generation method based on deep
  learning}.{\BBCQ}
\newblock
 \APACrefbtitle {Computational Science -- ICCS 2020} {Computational science --
  iccs 2020}\ (\BPGS\ 186--198).
\newblock
\APACaddressPublisher{Cham}{Springer International Publishing}.
\PrintBackRefs{\CurrentBib}

\bibitem [\protect \citeauthoryear {%
Zhou%
, Ren%
, Yang%
, Fan%
\BCBL {}\ \BBA {} Huang%
}{%
Zhou%
\ \protect \BOthers {.}}{%
{\protect \APACyear {2021}}%
}]{%
zhou2021sliding}
\APACinsertmetastar {%
zhou2021sliding}%
\begin{APACrefauthors}%
Zhou, H.%
, Ren, D.%
, Yang, X.%
, Fan, M.%
\BCBL {} Huang, H.%
\end{APACrefauthors}%
\unskip\
\newblock
\APACrefYearMonthDay{2021}{}{}.
\newblock
{\BBOQ}\APACrefatitle {Sliding Sequential CVAE with Time Variant Socially-aware
  Rethinking for Trajectory Prediction} {Sliding sequential cvae with time
  variant socially-aware rethinking for trajectory prediction}.{\BBCQ}
\newblock
\APACjournalVolNumPages{arXiv preprint arXiv:2110.15016}{}{}{}.
\newblock

\newblock

\PrintBackRefs{\CurrentBib}

\end{thebibliography}


\end{document}